\documentclass[acmsmall]{acmart}

\AtBeginDocument{%
  \providecommand\BibTeX{{%
    \normalfont B\kern-0.5em{\scshape i\kern-0.25em b}\kern-0.8em\TeX}}}

\setcopyright{acmcopyright}

\acmJournal{JETC}

\usepackage{amsmath,amsfonts}
\usepackage[caption=false]{subfig}
\usepackage{threeparttable}
\usepackage{multirow}
\usepackage[redeflists]{IEEEtrantools}
\usepackage{tikz}

\newcommand{\ineq}[1]{\footnotesize$#1$\normalsize}{}
\newcommand{\tech}{NCRTM}{}
\newcommand{\rev}[1]{\textcolor{black}{#1}}
\newcommand{\nrev}[1]{\textcolor{black}{#1}}
\newcommand{\mr}[1]{\textcolor{black}{#1}}
\newcommand{\minorrev}[1]{\textcolor{black}{#1}}

\begin{document}
\bstctlcite{IEEEexample:BSTcontrol}
\title{Dynamic Reliability Management in Neuromorphic Computing}

\author{Shihao Song}
\email{shihao.song@drexel.edu}
\author{Jui Hanamshet}
\email{jh3454@drexel.edu}
\author{Adarsha Balaji}
\email{ab3586@drexel.edu}
\author{Anup Das}
\email{anup.das@drexel.edu}
\affiliation{%
  \institution{Drexel University}
  \streetaddress{3101 Market Street}
  \city{Philadelphia}
  \state{PA}
  \postcode{19104}
}

\author{Jeffrey L. Krichmar}
\email{jkrichma@uci.edu}
\author{Nikil D. Dutt}
\email{dutt@ics.uci.edu}
\affiliation{%
  \institution{University of California, Irvine}
  \streetaddress{6210 Donald Bren Hall}
  \city{Irvine}
  \state{CA}
  \postcode{92697}
}

\author{Nagarajan Kandasamy}
\email{nk78@drexel.edu}
\affiliation{%
  \institution{Drexel University}
  \streetaddress{3101 Market Street}
  \city{Philadelphia}
  \state{PA}
  \postcode{19104}
}

\author{Francky Catthoor}
\email{Francky.Catthoor@imec.be}
\affiliation{%
  \institution{Imec}
  \streetaddress{Kapeldreef 75}
  \city{3001 Leuven}
  \country{Belgium}
}








\renewcommand{\shortauthors}{Song, et al.}

\begin{abstract}
Neuromorphic computing systems execute machine learning tasks designed with Spiking Neural Networks (SNNs).
These systems
are embracing non-volatile memory (NVM) to implement high-density and low-energy synaptic storage.
Elevated voltages and currents needed to operate NVMs cause aging of CMOS-based transistors in each neuron and synapse circuit in the hardware, drifting the transistor's parameters from their nominal values.
If these circuits are used continuously for too long, the parameter drifts cannot be reversed, resulting in permanent degradation of circuit performance over time, eventually leading to hardware faults. 
Aggressive device scaling increases power density and temperature, which further accelerates the aging, challenging the reliable operation of neuromorphic systems.
Existing reliability-oriented techniques periodically de-stress all neuron and synapse circuits in the hardware at fixed intervals, assuming worst-case operating conditions, without actually tracking their aging at run time. To de-stress these circuits, normal operation must be interrupted, which 
introduces
latency in spike generation and propagation, impacting
the inter-spike interval and hence, performance, e.g., accuracy.
We observe that in contrast to long-term aging, which permanently damages the hardware, short-term aging in scaled CMOS transistors is mostly due to Bias Temperature Instability (BTI). The latter is heavily workload-dependent and more importantly, partially reversible.
We propose a new architectural technique to mitigate the aging-related reliability problems in neuromorphic systems, by designing an intelligent run-time manager (\textbf{\tech{}}), which dynamically de-stresses neuron and synapse circuits in response to the short-term aging in their CMOS transistors during the execution of machine learning workloads, with the objective of meeting a reliability target. 
\tech{} de-stresses these circuits only when it is absolutely necessary to do so, otherwise reducing the performance impact by scheduling de-stress operations off the critical path. We evaluate \tech{} with state-of-the-art machine learning workloads on a neuromorphic hardware. Our results demonstrate that \tech{} significantly improves the reliability of neuromorphic hardware, with marginal impact on performance.
\end{abstract}

\ccsdesc[500]{Hardware~Neural systems}
\ccsdesc[500]{Hardware~Bio-embedded electronics}
\ccsdesc[500]{Hardware~Aging of circuits and systems}
\ccsdesc[500]{Software and its engineering~Runtime environments}

\setcopyright{acmcopyright}
\acmJournal{JETC}
\acmYear{2021} \acmVolume{1} \acmNumber{1} \acmArticle{1} \acmMonth{1} \acmPrice{15.00}\acmDOI{10.1145/3462330}

\keywords{Neuromorphic Computing, Machine Learning, Spiking Neural Network (SNN), Bias Temperature Instability (BTI), Lifetime Reliability, Non-Volatile Memory (NVM), Phase-Change Memory (PCM), Run-time Manager (RTM).}

\maketitle

\section{Introduction}\label{sec:introduction}
\begin{figure}[h]
\centering
\includegraphics[width=0.5\columnwidth]{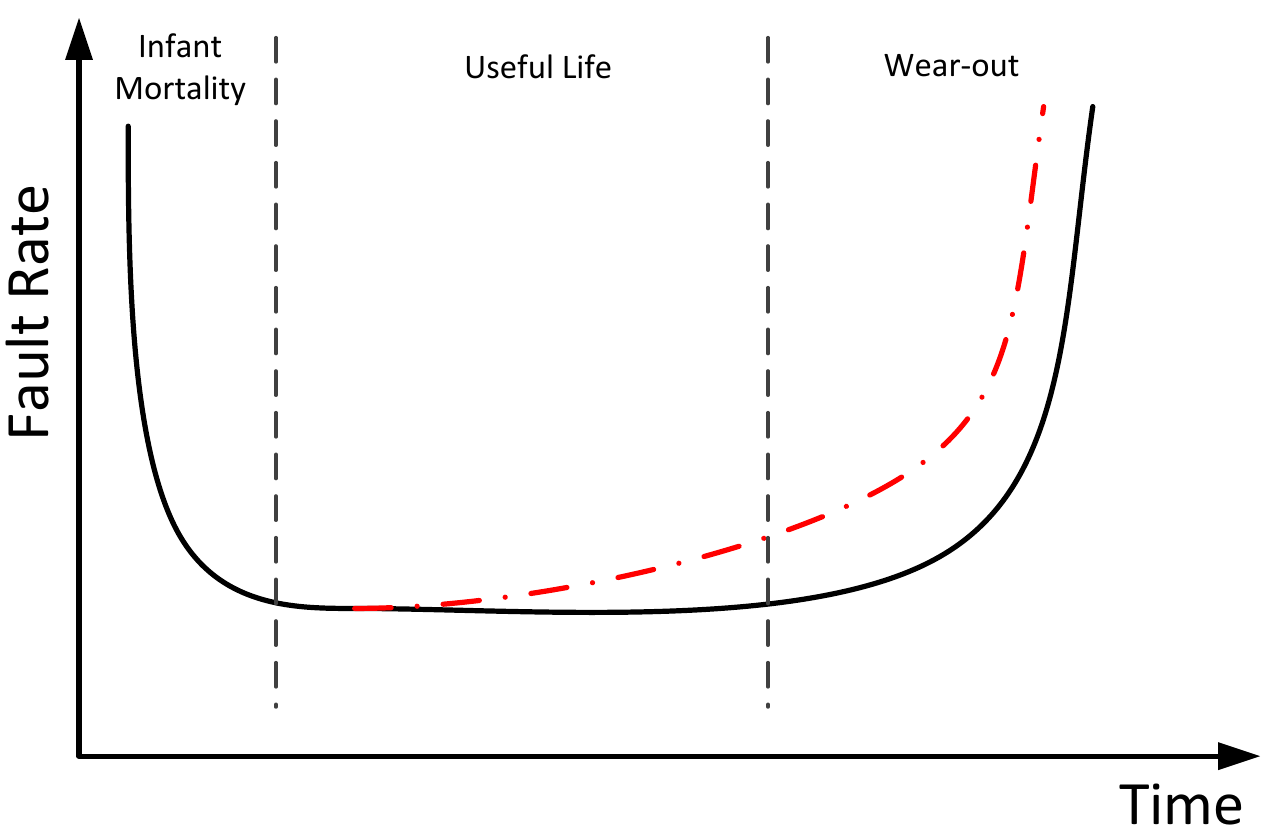}
\caption{Bathtub curve for permanent faults.}
\label{fig:bathtub}
\end{figure}

Spiking Neural Networks (SNNs)~\cite{maass1997networks} are machine learning approaches designed with spike-based computations~\cite{izhikevich2006polychronization} and bio-inspired learning algorithms~\cite{caporale2008spike} (See Appendix~\ref{sec:snns} for background on SNNs).
SNN-based workloads are typically executed on event-driven neuromorphic hardware such as TrueNorth~\cite{Debole2019}, Loihi~\cite{DaviesLoihi2018}, and DYNAP-SE~\cite{moradiDynap2017}. These hardware platforms are extremely energy-efficient, thanks to their event-driven activation
and their tile-based distributed architecture with in-place neural computations and synaptic storage~\cite{rajendran2019low}.
We investigate the internal architecture of neurons and synapses in DYNAP-SE (see Figures~\ref{fig:crossbar}b and \ref{fig:neuron}), and found that 
these circuits consist of transistors built using bulk CMOS or FinFet technologies~\cite{chen20184096,hisamoto2000finfet,azghadi2016hybrid}.\footnote{We believe these architectures are similar for other designs like TrueNorth~\cite{Debole2019} and Loihi~\cite{DaviesLoihi2018}.}
When operated at a high voltage and temperature, the transistor's parameters strongly drift from their nominal values. This is called \emph{aging}. In fact, in scaled technology nodes, this aging happens even under nominal conditions and from the very start of using the devices leading to the so-called soft breakdown. The most important breakdown mechanism is the Bias Temperature Instability (BTI)~\cite{weckx2014non,kraak2019parametric,kraak2018degradation}. Strongly depending on the workload, BTI is highly variable and it is largely reversible under nominal conditions on removal of the stress voltage. So it leads only to parametric time-dependent variability, affecting mainly delay and leakage power. If the neurons and synapses in a neuromorphic hardware are used continuously for long duration at elevated operating conditions, the parameter drifts cannot be reversed~\cite{6531944}, leading to permanent functional degradation of the circuit and eventually, hardware faults~\cite{pae2015considering,taghipour2017aging,kukner2014degradation}. 
\mr{
The permanent fault rates in integrated circuits can be described by the bathtub curve as shown in Figure~\ref{fig:bathtub}. 
Post manufacturing, integrated circuits (IC) are characterized by high failure rates as these circuits are subjected to manufacturing tests, such as stuck-at, at-speed, burn-in, etc., which filters out defective circuits and circuits with short lifetime. The probability of the successful circuits surviving for a longer period of time, increases. The failure rate, therefore, decreases over time. This phase is known as the infant mortality period. This is followed by a period of constant failure rate, often referred as useful life. The last phase is known as the wear-out or the aging phase and is characterized by increasing fault rate.  Recent studies on reliability reveal that, if wear-out is not addressed from early device usage stage (e.g., the beginning of useful life period), circuits can age faster than anticipated with the wear-out phase settling earlier in life (shown by the red dashed line in the figure).
}

\mr{
To address time-dependent variability or aging, circuit designers often set 
worst-case and hence highly pessimistic
reliability-related extra design margins, which unnecessarily constrain performance. Our objective is to analyze the circuit aging in neuromorphic hardware at real-time and take corrective measures at the architecture-level to reverse the parameter drifts based on the utilization of neuron and synapse circuits within a machine learning workload.
}

Recently, Non-Volatile Memory (NVM) is used in neuromorphic hardware to implement high-density and low-energy synaptic storage~\cite{Burr2017}. Several NVMs are explored for this purpose -- Oxide-based Resistive RAM (OxRRAM)~\cite{Mallik2017}, Phase Change Memory (PCM)~\cite{nandakumar2018phase}, Ferro-Electric RAM~\cite{mulaosmanovic2017novel}, and Spin-Transfer Torque Magnetic or Spin-Orbit-Torque RAM (STT- and SoT-MRAM)~\cite{vincent2015spin}.\footnote{Beside neuromorphic computing, NVMs are also used as main memory for conventional computing~\cite{palp,datacon,mneme,hebe,kultursay2013evaluating,lee2009architecting,qureshi2009scalable}.}
NVMs require either high voltage (OxRRAM, PCM and FeFET) or high current (MRAM) to operate, which accelerates the aging of transistors in neuron and synapse circuits in a neuromorphic hardware~\cite{frameworkCAL,reneu,song2020case,vts_das,shihao_igsc}.
Aggressive device scaling increases power density and temperature, which makes reliability even worse. Therefore, circuit aging is emerging as one of the primary reliability concerns for neuromorphic hardware designed with NVMs~\cite{cartier2019reliability}. 

\mr{
The reliability problem we are addressing in this work is due to high voltage operations of NVMs. That can also occur in other system contexts,\footnote{\mr{BTI issues are also a reliability concern for standard DRAM and SRAM memories~\cite{navarro2019reliability}. However, due to the use of transistors as access devices, the peripheral circuits in DRAM and SRAM can use lower operating voltages $\approx$ 1.2V. BTI-related reliability issues in DRAM and SRAM are therefore less severe than in NVM contexts~\cite{mneme,hebe}.}}  but it is in particular an issue for SNNs due to the following reasons.
To address this high voltage NVM problem, periodic de-stress of the peripheral circuit is necessary, which impacts inter-spike interval (ISI) when machine learning models are executed on these circuits. 
The performance (e.g., accuracy) of SNNs depends on ISI.
Therefore, the reliability issues of NVMs lead to performance issues in SNNs.
}

Prior works on mapping machine learning workloads to neuromorphic hardware have mostly focused on compilation techniques, with the objective of improving machine learning performance on hardware.
\minorrev{
Examples of such approaches include
hardware utilization-based mapping~\cite{ji2016neutrams,esl20,galluppi2015framework,dfsynthesizer,rtmJSPS,pycarl,ji2018bridge,adarsha_igsc,balaji2019ISVLSIframework,das2018dataflow}, energy-based mapping~\cite{spinemap,psopart,twisha_energy}, and endurance-based mapping~\cite{twisha_endurance,twisha_thermal,espine}.
}
The recently-proposed approach RENEU~\cite{reneu} is the only compile-time based technique that 
maps the 
neurons and synapses to the hardware to improve the long-term, i.e., the lifetime reliability. 
\mr{
Although compile-time based aging mitigation approaches have unique advantages such as low computation overhead, predictability, and performance guarantee, they are often conservative and therefore, may miss significant performance and reliability improvement opportunities. Dynamic approaches are flexible, adaptive, and potentially more effective in a highly dynamic environment, such as ones where the inference data deviates strongly from training examples.
}
We show that both performance and reliability can be improved significantly if neuron and synapse circuits are de-stressed periodically at run-time based on current data.

On the run-time front, very few approaches address the run-time management of neuromorphic computing.\footnote{There are works that address run-time management for conventional multi-core systems~\cite{shafik2015adaptive}.} In~\cite{rtmJSPS}, the authors propose a fast approach to remap online learning SNNs on a neuromorphic hardware after every learning epoch to improve model performance. In DTRO~\cite{frameworkCAL}, the authors propose a hybrid approach to estimate the reliability degradation for machine learning workloads at design-time using training data, and use this information to de-stress all hardware circuits during run-time at fixed intervals, without actually tracking the circuit aging.
\rev{
The effectiveness of this approach is limited to supervised techniques only and the availability of representative training data.
To this end, we make the following three key observations. 
}
\begin{itemize}
    \item \textbf{Observation 1:} \emph{\mr{Workload, which includes synaptic weights and their activation on neuromorphic hardware, is specific to the machine learning task being executed and its input.}}
    \item \textbf{Observation 2:} \emph{\mr{De-stressing all circuits in the hardware periodically, without tracking the actual aging, introduces long latency in spike generation and propagation, which impacts inter-spike interval, leading to information loss in SNNs.}}
    \item \textbf{Observation 3:} \emph{Compared to long-term aging under elevated stress conditions, which is permanent and irreversible, short-term aging under nominal conditions is heavily workload-dependent (and hence to some extent controllable), and partially reversible.}
\end{itemize}

Based on these three observations, we introduce \tech{}, a run-time reliability manager for neuromorphic hardware to de-stress neuron and synapse circuits in the hardware only when needed, by dynamically tracking their short-term aging during the execution of machine learning tasks.
\tech{} extends our earlier work DTRO~\cite{frameworkCAL} with the following \textit{new} contributions.

\begin{itemize}
    \item We introduce an intelligent run-time manager \textbf{\tech{}}, which improves the long-term reliability of neuromorphic hardware by controlling its short-term aging when executing machine learning tasks.
    \item We develop a run-time performance monitoring and reliability estimation framework using statistics collected from the neuromorphic hardware.
    \item We show that \tech{} can be applied to both supervised and unsupervised machine learning approaches and scenarios where the number of training examples are limited.
    \item We evaluate \tech{} with machine learning workloads designed using Convolution Neural Network (CNN), Multi-layer Perceptron (MLP), and Recurrent Neural Network (RNN) models on a state-of-the-art neuromorphic hardware simulator.
\end{itemize}

Overall, \tech{} mitigates the aging-related reliability problems in neuromorphic computing by dynamically de-stressing neuron and synapse circuits in response to their short-term aging, with the objective of meeting a reliability target. 
\nrev{
\tech{} de-stresses these circuits only when it is absolutely necessary to do so, otherwise reducing the performance impact by scheduling all de-stress operations off the critical path by tracking the latency impact of de-stress operations on inter-spike interval (ISI), a key performance measure in SNNs.
}

\section{Comparison with State-of-the-art}\label{sec:sota}
\begin{figure}[h!]%
    \centering
    \subfloat[Design-time approach (e.g., RENEU~\cite{reneu}).]{{\includegraphics[width=4.3cm]{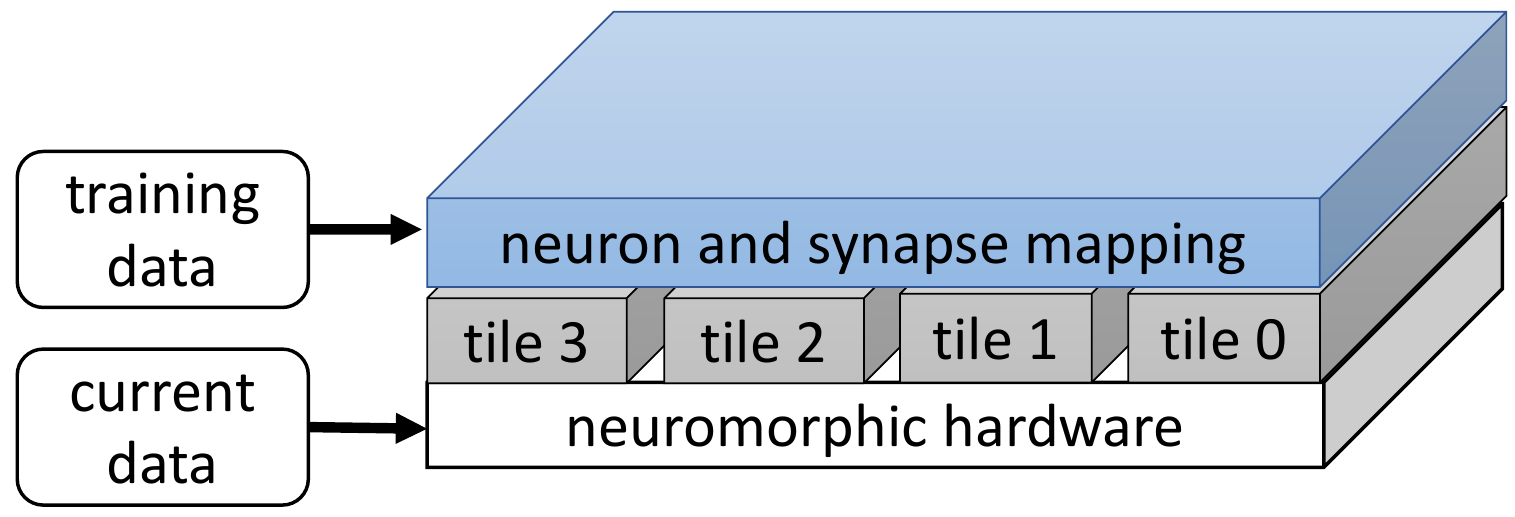} }}%
    \quad
    \subfloat[Hybrid approach (e.g., DTRO~\cite{frameworkCAL}).]{{\includegraphics[width=3.5cm]{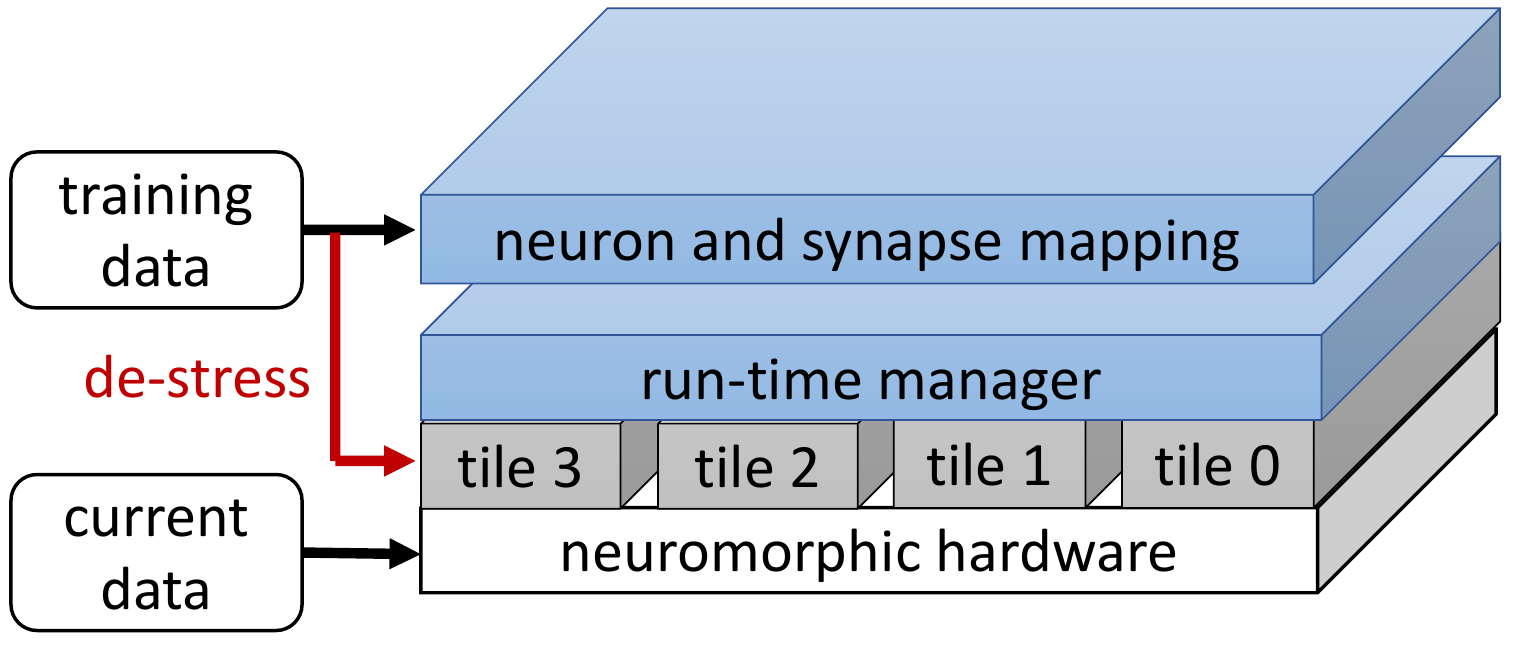} }}%
    \quad
    \subfloat[Run-time approach (proposed, \tech{}).]{{\includegraphics[width=4cm]{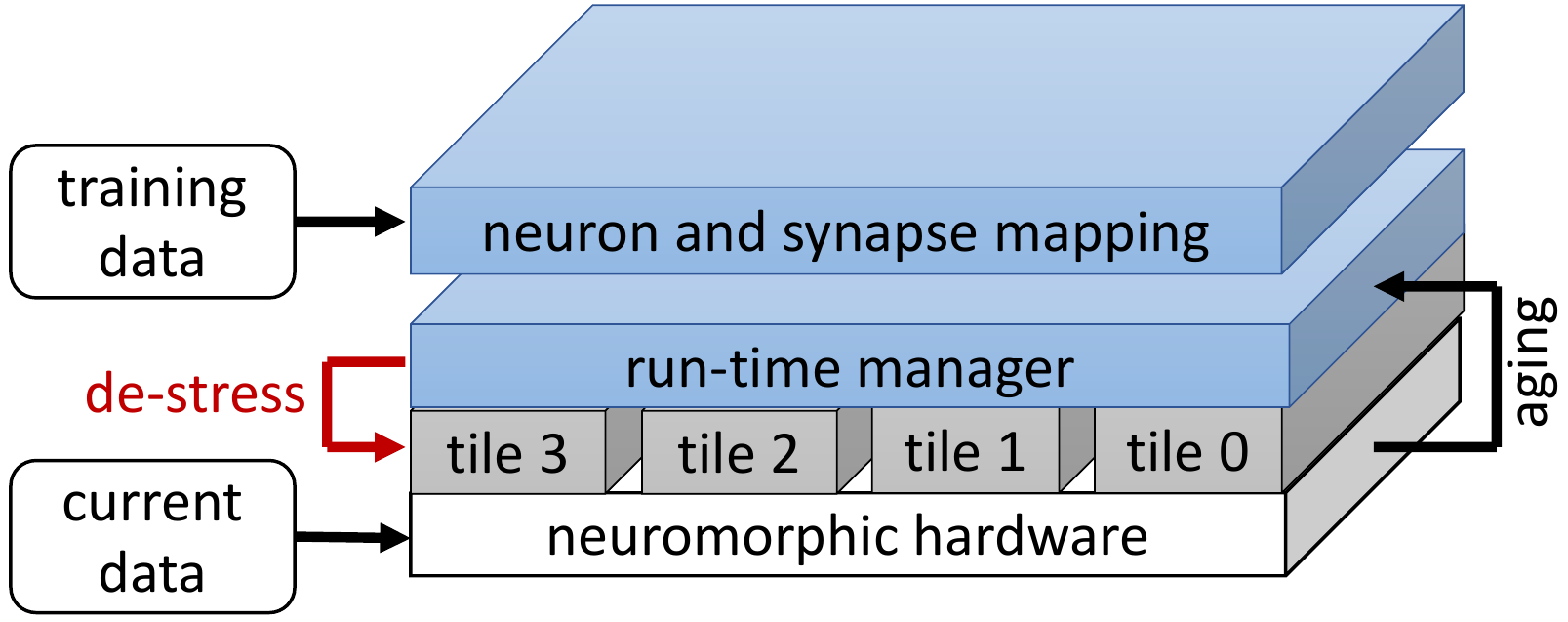} }}%
    \caption{Illustrating the trade-off between neuron aging and SNN performance.}%
    \label{fig:sota}%
\end{figure}

Figure~\ref{fig:sota} illustrates how the proposed approach differs from two reliability-oriented state-of-the-art approaches.
Figure~\ref{fig:sota}a illustrates a design-time approach such as RENEU~\cite{reneu}, where neurons and synapses are mapped to the hardware to increase the long term, i.e., the lifetime reliability. This approach
estimates the aging in neuron and synapse circuits using representative training examples. There are no corrective online measures in place to control the aging, should the aging exceed a critical threshold or the workload behavior changes, for instance, when encountering unseen data at run-time.
Figure~\ref{fig:sota}b illustrates a hybrid approach such as DTRO~\cite{frameworkCAL}, where neurons and synapses are mapped to the hardware using a reliability-oriented mapping technique (e.g., RENEU~\cite{reneu}). Additionally, all neuron and synapse circuits in the hardware are periodically de-stressed to control the aging. The de-stress interval is determined using training examples. The drawbacks of such an approach are the following. First, by not tracking the actual aging in real-time, such approach can introduce significant latency in interrupting normal operation, even when the aging is much below the critical threshold. This is especially critical for SNNs because the performance of machine learning workloads, e.g., their accuracy, depends on the precise times of spikes (see Appendix~\ref{sec:snns}). Second, the effectiveness of such a hybrid approach depends heavily on the training data, which may not always be representative. In fact, hybrid approaches present significant limitations for unsupervised applications or applications with limited training data.

\rev{
Figure~\ref{fig:sota}c illustrates \tech{}, the proposed run-time approach for reliability management in neuromorphic hardware. \tech{} tracks the aging in neuron and synapse circuits at real-time during the execution of machine learning workloads and de-stresses these circuits only when their aging exceeds a critical threshold. By implementing age tracking and control at run-time, the de-stress decisions of \tech{} are made based on current data. Therefore, \tech{} is relevant for both supervised and unsupervised machine learning approaches. 
}

\rev{
To the best of our knowledge, \tech{} is the first work for run-time reliability management of neuromorphic hardware. In Section~\ref{sec:results}, we evaluate \tech{} against these state-of-the-art reliability management approaches using both supervised and unsupervised machine learning workloads. 
}

\section{Background}\label{sec:background}
In this section, we introduce the background necessary to understand our proposed run-time manager \tech{}. Background on SNNs are provided as Appendix.
\subsection{Neuromorphic Hardware}
\begin{figure}[h!]
	\centering
	\centerline{\includegraphics[width=0.69\columnwidth]{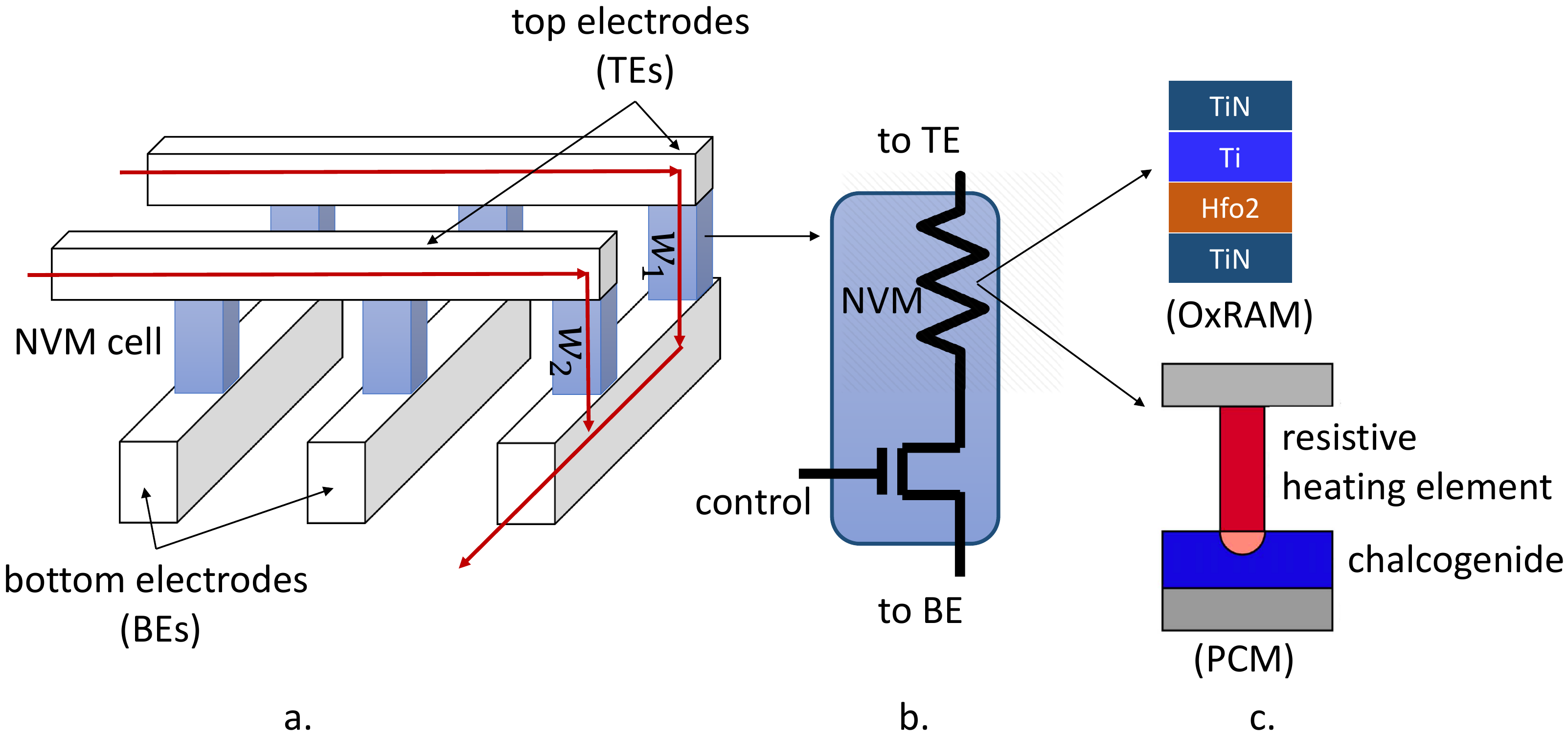}}
	\caption{(a) A crossbar of a neuromorphic hardware, (b) a synaptic cell consisting of a NVM device (a resistive element) and a transistor, and (c) Hfo2-based OxRAM and chalcogenide-based PCM as NVM device.}
	\label{fig:crossbar}
\end{figure}

\minorrev{
We consider tile-based neuromorphic hardware~\cite{catthoor2018very,balaji2019design}, where the tiles are interconnected using networks-on-chip (NoC) or Segmented Bus~\cite{balaji2019exploration}. 
This is
similar to
several contemporary neuromorphic architectures such as
DYNAP-SE~\cite{moradiDynap2017}, Loihi~\cite{DaviesLoihi2018}, and TrueNorth~\cite{Debole2019}.
}
Each tile in the hardware consists of a crossbar for synaptic storage, a set of input and output neurons, and a performance monitoring unit, which in its simplest form is a \textit{spike counter} (SC). A crossbar, shown in Figure~\ref{fig:crossbar}a, is a 3D organization of top electrodes (TEs), which form the rows and bottom electrodes (BEs), which form the columns. 
A synaptic cell is connected at a crosspoint, i.e., at the intersection of each row and column via an access transistor as shown in Figure~\ref{fig:crossbar}b.
Pre-synaptic neurons are mapped along the TEs and post-synaptic neurons along the BEs. The synaptic weight between a pre- and a post-synaptic neuron is programmed as conductance of the corresponding synaptic cell at the crosspoint. \mr{A pre-synaptic neuron's voltage (\ineq{V_i}) applied on the TE is multiplied by the conductance (\ineq{G_i}) to generate current \ineq{I_i = V_i\cdot G_i} (according to Ohm's Law). This current propagates to the post-synaptic neuron to raise its action potential. Current summation occurs on each BE according to Kirchoff's Current Law, when integrating excitation from other pre-synaptic neurons. This implements \ineq{\sum_i I_i = \sum_i V_i\cdot G_i}. This is the in-memory multiply accumulate logic implemented inside a crossbar.}  Figure~\ref{fig:crossbar}a illustrates the integration of input excitation from two pre-synaptic neurons to one post-synaptic neuron via the synaptic weights \ineq{w_1} and \ineq{w_2}, respectively. This forms the \emph{data plane} of the neuromorphic hardware. The \textit{control plane} of the hardware consists of control signals, which enable specific access transistors (see Figure~\ref{fig:crossbar}b) to facilitate current flow in the crossbar. 
The NVM device of a synaptic cell, shown as a resistive element in Figure~\ref{fig:crossbar}b, can be implemented for instance with HfO2-based OxRAM or chalcogenide-based PCM as shown in Figure~\ref{fig:crossbar}c. But our approach is not limited to these specific NVM technologies.

\begin{figure}[h]%
	\centering
	\subfloat[][Current and voltages between the pre-synaptic neuron \ineq{N_i} and the post-synaptic neuron \ineq{N_j}. \label{fig:current_voltage}]{
		\includegraphics[width=0.3\columnwidth]{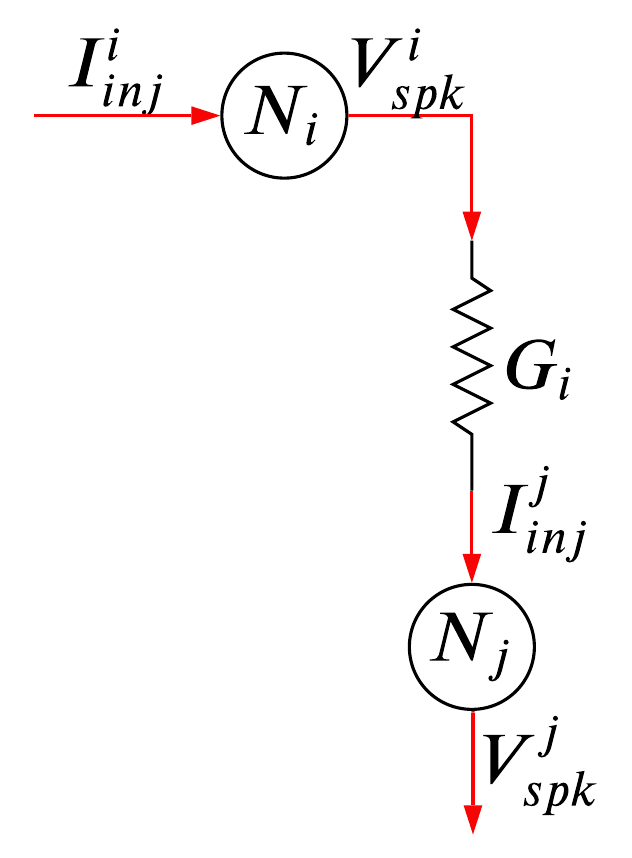}
	}
	\quad
	\subfloat[][Internal architecture of a leaky integrate and fire (LIF) neuron~\cite{indiveri2003low}, showing different transistors and reference voltages needed to generate the necessary spike voltage at the output. \label{fig:neuron}]{
		\includegraphics[width=0.6\columnwidth]{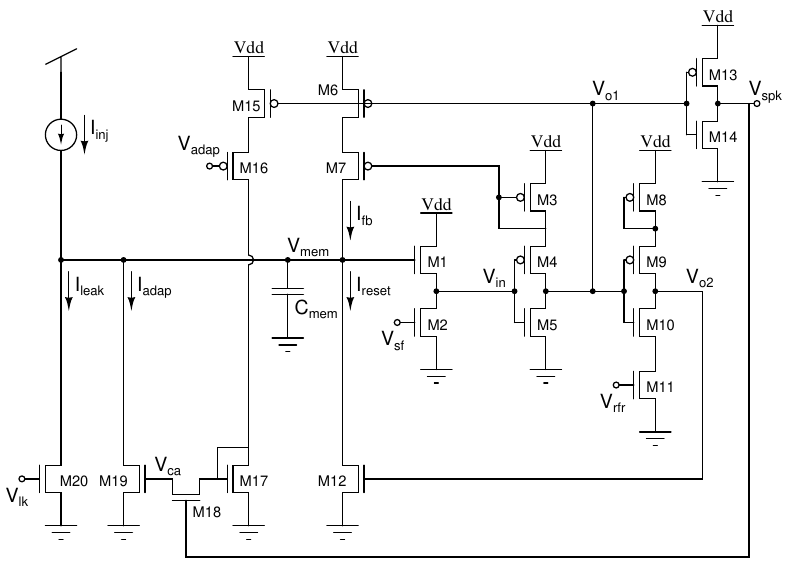}
	}
	\caption{a) Connection between neurons and b) a leaky integrate and fire (LIF) neuron~\cite{indiveri2003low}.}
	\label{fig:neuron_all}
\end{figure}

\mr{
Figure~\ref{fig:current_voltage} shows the currents and voltages on the path from the pre-synaptic neuron \ineq{N_i} to the post-synaptic neuron \ineq{N_j}. The input current \ineq{I_{inj}^i} is converted into voltage \ineq{V_{spk}^i} using the neuron \ineq{N_i}. This voltage is multiplied with the conductance \ineq{G_i} (representing synaptic weight) to generate the current \ineq{I_{inj}^j}. This current is converted to voltage \ineq{V_{spk}^j} using neuron \ineq{N_j}.
}
Figure~\ref{fig:neuron} illustrate the internal architecture of a Leaky-Integrate-and-Fire neuron~\cite{indiveri2003low}. The current \ineq{I_\text{inj}} injected into the neuron is proportional to the weighted sum of excitation from all of its pre-synaptic connections. The PMOS and NMOS transistors in the neuron and the reference voltages raise the neuron's membrane voltage. When the voltage crosses a threshold, a spike is generated. The spike voltage (\ineq{V_\text{spk}}) must be sufficiently high to propagate current through the synaptic cell connected at the output of the neuron.

Certain NVMs such as PCM, OxRRAM, and FeFET requires high voltages to operate. 
We consider the case of PCM-based neuromorphic hardware, which requires \ineq{\approx} 3V~\cite{mneme} to propagate current through it. This high voltage causes aging of the access transistor in each synaptic cell in a crossbar and also of the transistors in each neuron connected along the TEs and the BEs of the crossbar.


\subsection{Transistor Aging in Neuromorphic Hardware}
High voltage operations of transistors introduce many reliability issues such as Time-Dependent Dielectric Breakdown (TDDB), Bias Temperature Instability (BTI), and Hot-Carrier Injection (HCI). 
These are the dominant causes of aging in scaled technology nodes (45nm and below)~\cite{mahmoud2019comparative}. In older nodes, Electromigration (EM) also plays a key role~\cite{pierce1997electromigration,das2013communication,santos2014criticality,das2014communication,das2015workload,das2012fault,das2015reliability,das2014temperature,das2014reinforcement}. 

Transistor aging is accelerated when it is \emph{stressed}, i.e., exposed to high overdrive voltage, where overdrive voltage is defined as the voltage between transistor gate and source ($V_{GS}$) in excess of the threshold voltage ($V_\text{th}$), where $V_\text{th}$ is the minimum voltage required between gate and source to turn the transistor on. With this understanding, 
we provide a brief background of these three failure mechanisms.

\begin{itemize}
    \item \textit{TDDB:} This is a failure mechanism in a CMOS device, when the gate oxide breaks down as a result of long-time application of relatively low electric field (as opposed to immediate breakdown, which is caused by strong electric field)~\cite{roussel2018new}. The lifetime of a CMOS device is measured in terms of its \textit{mean time to failure} (\textbf{MTTF}) as
\begin{equation}
    \label{eq:MTTF_TDDB}
    \footnotesize \text{MTTF}_\text{TDDB} = A.e^{-\gamma\sqrt{V}},
\end{equation}
where \ineq{A} and \ineq{\gamma} are material-related constants, and \ineq{V} is the overdrive gate voltage of the CMOS device. 
\item \emph{BTI:} This is a failure mechanism in a CMOS device where positive charges are trapped at the oxide-semiconductor boundary underneath the gate~\cite{gao2017nbti}.
BTI manifests as 1) decrease in drain current and transconductance, and 2) increase in off current and threshold voltage. 
The BTI lifetime of the device is
\begin{equation}
    \label{eq:MTTF_NBTI}
    \footnotesize \text{MTTF}_\text{BTI} = \frac{A}{V^\gamma}e^{\frac{E_a}{KT}},
\end{equation}
where \ineq{A} and \ineq{\gamma} are material-related constants, \ineq{E_a} is the activation energy, \ineq{K} is the Boltzmann constant, \ineq{T} is the temperature, and \ineq{V} is the overdrive gate voltage of the CMOS device.
\item \emph{HCI:} This is a failure mechanism in a CMOS device, when a carrier (electron or hole) gains sufficient kinetic energy to overcome the potential barrier of the conducting channel and gets trapped in the gate dielectric, permanently changing its switching characteristic~\cite{wan2019hci}.
\end{itemize}

Unlike the TDDB and BTI failure mechanisms, for which silicon-characterized reliability models are available from foundries, characterized models for HCI failure mechanisms are still in development for scaled nodes.
\mr{
Among these failure mechanisms, BTI is generally accepted as the most important mechanism for sub-10 nm nodes~\cite{ramey2018technology,liu2020new,sagong2020reliability}.} HCI mostly occurs there under strong voltage/current conditions and TDDB has become less important because technologists have stopped pursuing ultra-high k values in the dielectric.

\section{Observations Leading to \tech{}}\label{sec:observations}
We expand on the three observations made in Section~\ref{sec:introduction}.

\subsection{Observation 1: Workload-dependent Activation} 
\begin{figure}[h!]
 	\centering
    \vspace{-6pt}
 	\centerline{\includegraphics[width=0.99\columnwidth]{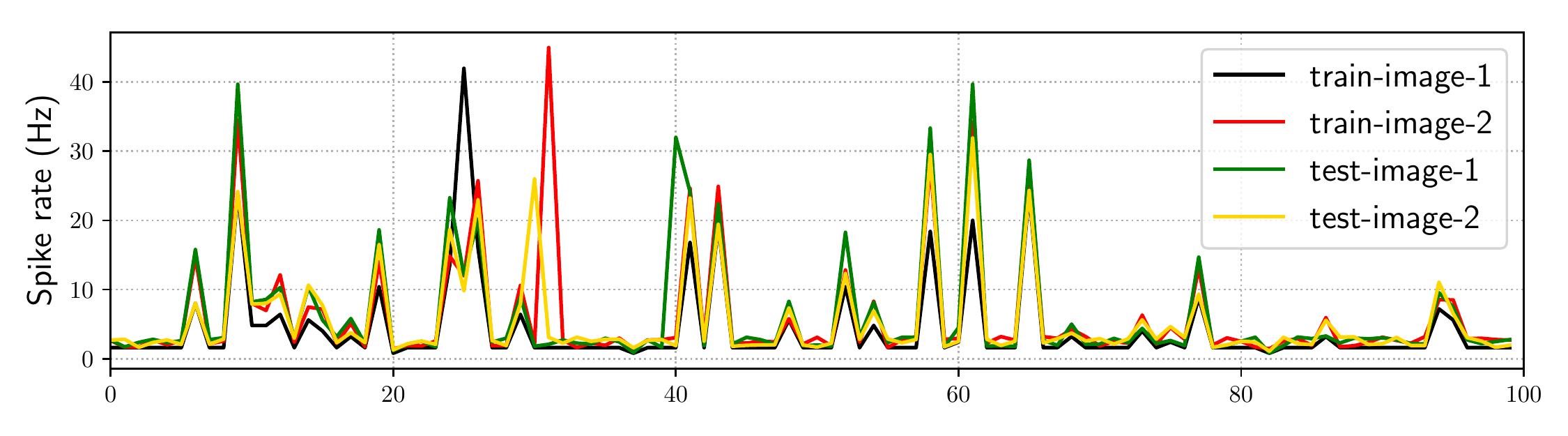}}
 	\vspace{-10pt}
 	\caption{Spike rate of 100 randomly-selected neurons in AlexNet for 2 training images and 2 test images.}
 	\label{fig:alexnet}
\end{figure}

To illustrate the application and input-dependent neuron activation, Figure~\ref{fig:alexnet} plots the spike firing rate of 100 randomly-selected neurons in AlexNet~\cite{alexnet}, a state-of-the-art CNN used for Imagenet classification. We report results for two randomly-selected training and test images.

We observe that spike firing rates of neurons depend on the image presented to the AlexNet CNN. Therefore, reliability improvement strategies based on design-time analysis with training examples may not be optimal when they are applied at run-time to process in-field data, a limitation of our prior work~\cite{frameworkCAL}. We address this limitation by designing our proposed run-time framework \tech{}, which can adapt its decisions based on current data.

\begin{figure}[h!]
 	\centering
    \vspace{-6pt}
 	\centerline{\includegraphics[width=0.99\columnwidth]{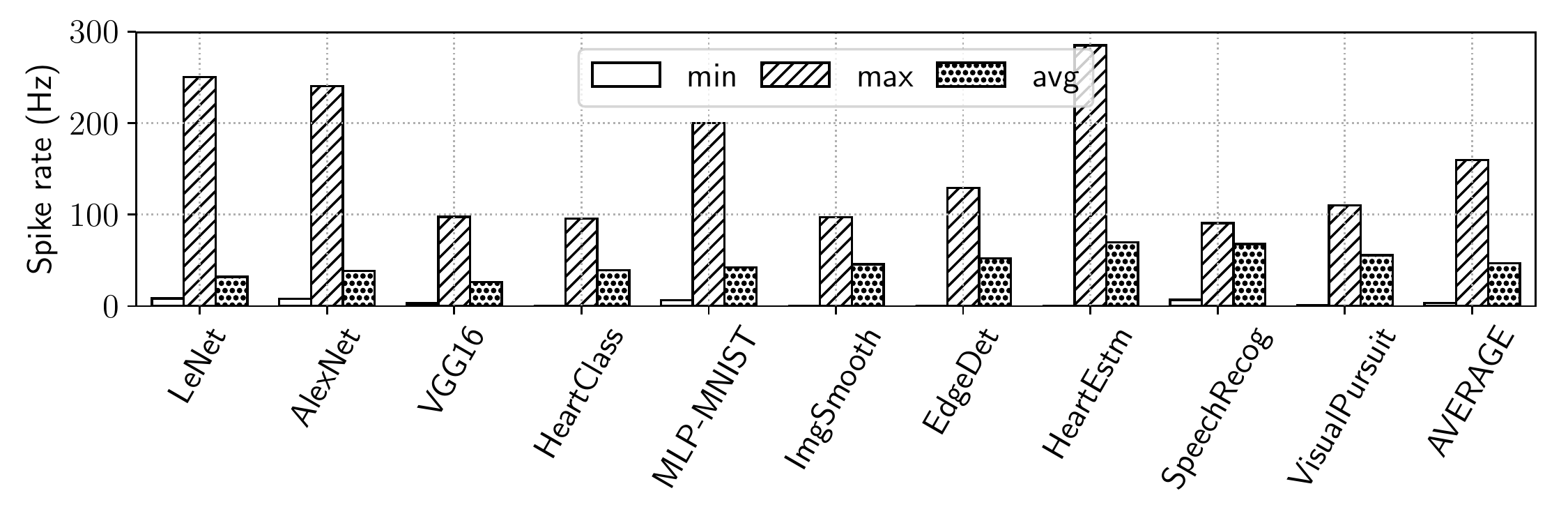}}
 	\vspace{-10pt}
 	\caption{Minimum, maximum, and average spike rate of all neurons in 10 workloads for test set.} 
 	\label{fig:workload}
\end{figure}

To demonstrate the workload-dependent nature of spike firing rate, Figure~\ref{fig:workload} plots the minimum, maximum, and average spike rate of all neurons in 10 machine learning workloads (see Section~\ref{sec:evaluation}) for test examples from their respective dataset. We observe that spike rates of neurons are strongly workload-dependent, and therefore, a workload-specific strategy is needed to optimally control the reliability aspect. This is precisely the objective of \tech{}.

\subsection{Observation 2: Performance Trade-off in Reliability Improvement}
\begin{figure}[h!]%
    \centering
    \subfloat[Example spike train from a neuron in AlexNet.]{{\includegraphics[width=13cm]{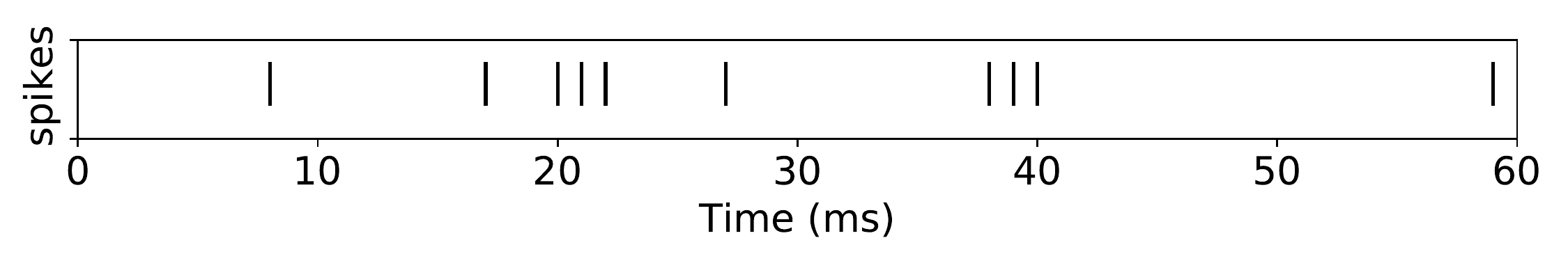} }}%
    \qquad
    \subfloat[Charge pump voltage to process the spike train.]{{\includegraphics[width=13cm]{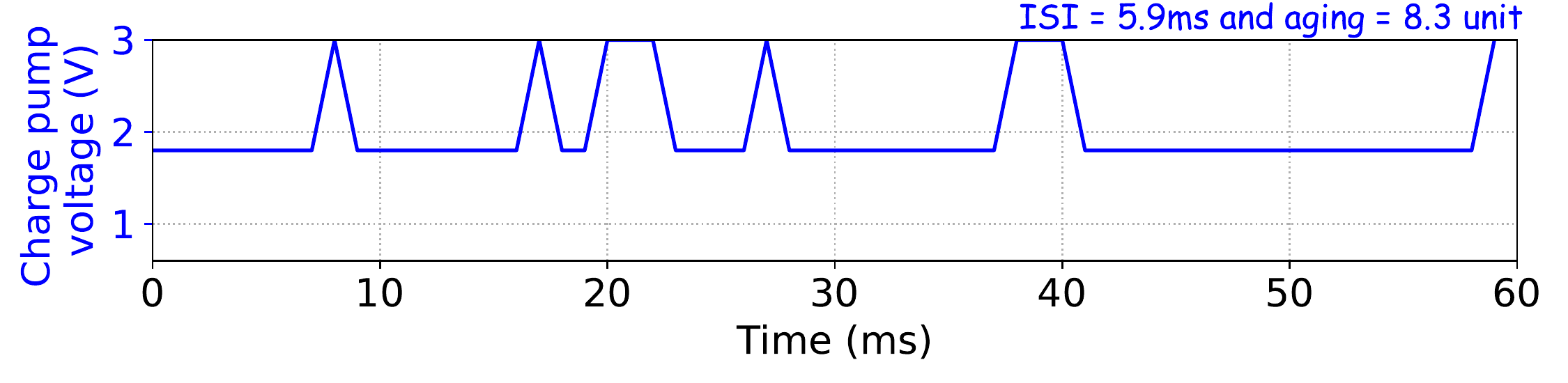} }}%
    \qquad
    \subfloat[Charge pump reset to 1.2V after processing every spike.]{{\includegraphics[width=13cm]{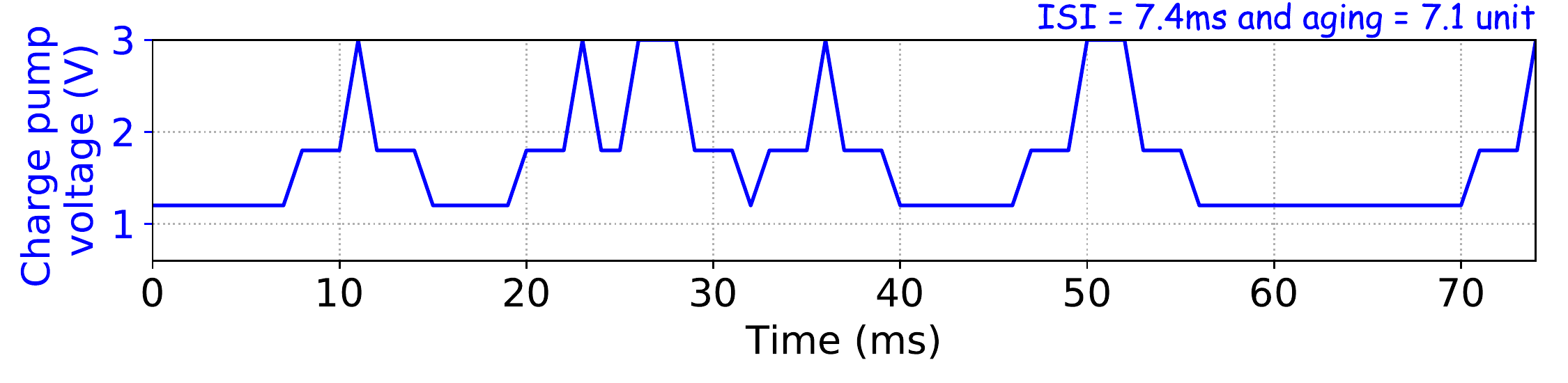} }}%
    \caption{Trade-off between neuron aging and SNN performance.}%
    \label{fig:cp_motivate}%
\end{figure}

In SNNs, information is encoded in spike times. Inter-spike interval (ISI) defines the performance metric in SNNs. To demonstrate how ISI is impacted by reliability-oriented decisions,
Figure \ref{fig:cp_motivate}(a) shows the spike train generated by a neuron in AlexNet when processing a reference image. Each spike injects current into the crossbar to flow through the NVM cell. Figure \ref{fig:cp_motivate}(b) illustrates the voltage of the on-chip charge pump that supply the reference voltages in the neuron to generate this spike train. The charge pump is operated at \ineq{1.8V} for the entire 60ms interval, boosting its voltage to \ineq{3V} only to generate spikes. 
Aging of the transistors in the neuron is 8.3 units (see Section \ref{sec:aging} 
for aging computation) and the average ISI is 5.9ms (See Section \ref{sec:isi} for ISI computation).
Figure \ref{fig:cp_motivate}(c) illustrates the charge pump's operating voltage when it is discharged to \ineq{1.2V} after generating every spike and boosted again to \ineq{1.8V} before generating the next. This is to de-stress the transistors in the neuron. 
Once de-stressed, the neuron becomes unavailable to generate spikes, introducing latency in the spike train. The average ISI increases to 7.4ms, compared to 5.9ms in Figure \ref{fig:cp_motivate}(b). {Changes in ISI may lead to accuracy loss}.
Frequently discharging the charge pump, however, reduces the neuron's aging to 7.1 units, compared to 8.3 units in Figure \ref{fig:cp_motivate}(b). This reduction in aging leads to an improvement of MTTF, i.e., the lifetime of the neuron. 

\mr{
Although it is possible to estimate the worst-case ISI degradation at compile-time using SpiNe\-Map~\cite{spinemap} and other similar approaches, such estimation can deviate significantly from the actual case in a highly dynamic environment, where testing data is different from the training examples (see Figure~\ref{fig:alexnet}). Therefore, a run-time manager is desirable to dynamically adapt the reliability-oriented decisions to limit ISI degradation.  
}

\subsection{Observation 3: Short-term vs. Long-term Aging}
In this work, we demonstrate our approach for BTI-related failures. 
The general principle applies to any other failure mechanism.
\mr{BTI aging manifests as: 1) A decrease in drain current and transconductance, and 2) An increase in off current and threshold voltage.  
When operated at a high voltage and temperature, these parameters strongly drift from their nominal values. In fact, in scaled technology nodes, this BTI aging happens even under nominal conditions and from the very start of using the devices leading to the so-called soft breakdown~\cite{weckx2014non,kraak2019parametric,kraak2018degradation,das2018reliable}.}

\begin{figure}[h!]
 	\centering
    \vspace{-6pt}
 	\centerline{\includegraphics[width=0.6\columnwidth]{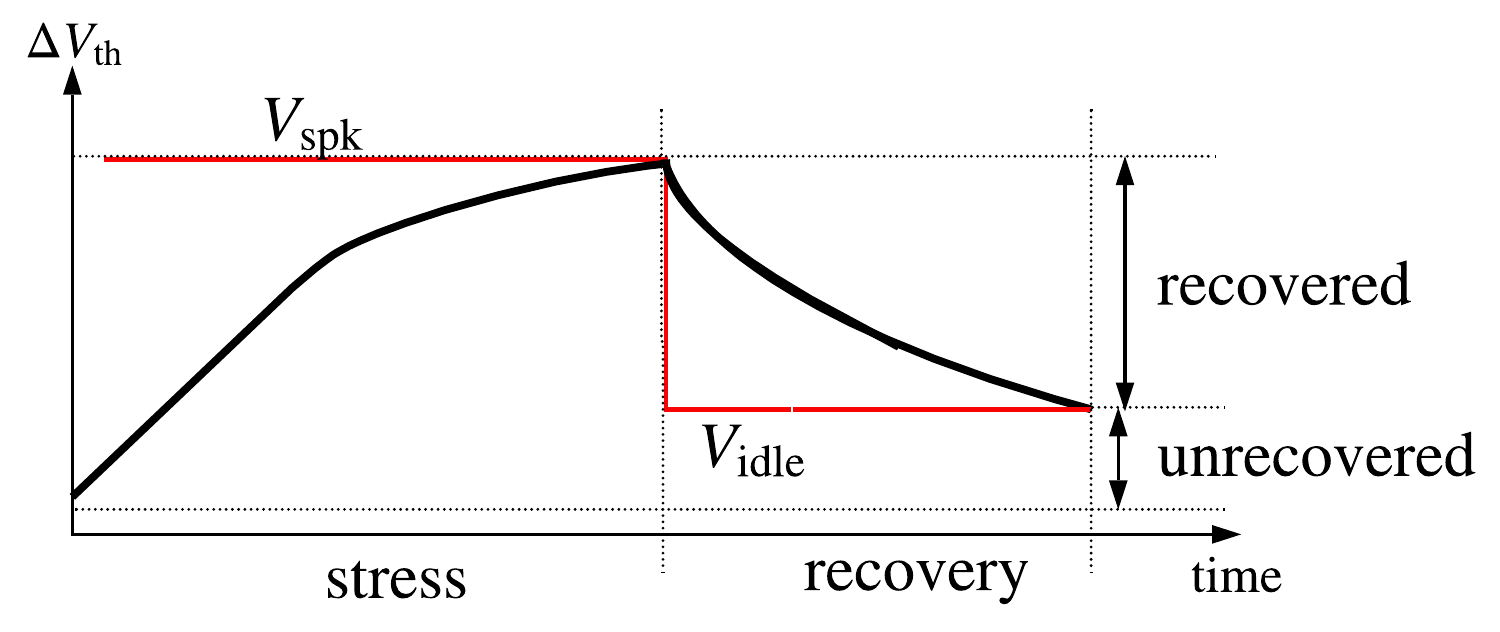}}
 	\vspace{-10pt}
 	\caption{Demonstration of degradation due to BTI.}
 	\label{fig:nbti_demo}
\end{figure}

Recent works such as~\cite{gao2017nbti,kraak2018degradation,kraak2019parametric,weckx2014non,puschkarsky2018understanding,rzepa2018comphy,grasser2017implications} suggest that BTI is the collective response of two independent defects -- the \textit{as-grown hole traps} (AHTs) and \textit{generated defects} (GDs). AHTs and a small proportion of GDs can be recovered by annealing at high temperatures if the BTI stress voltage is removed (\textit{de-stress}). 
Figure~\ref{fig:nbti_demo} illustrates the stress and recovery of the threshold voltage of a CMOS transistor on application of a high (\ineq{V_\text{spk}}) and a low voltage (\ineq{V_\text{idle}}). We observe that both stress and recovery depends on the time of exposure to the corresponding voltage level. This implies that when a neuron is idle, the BTI aging of the neuron recovers from stress.

\begin{figure}[h]%
	\centering
	\subfloat[][BTI aging when de-stressing a neuron circuit once every second. \label{fig:nbti_demo_1}]{
		\includegraphics[width=1.0\columnwidth]{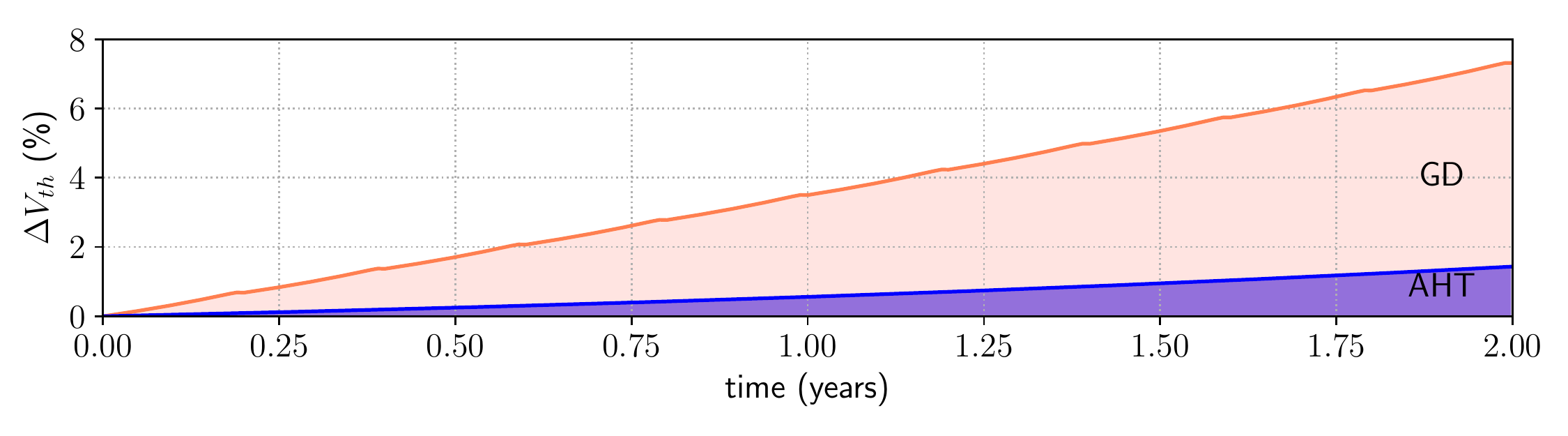}
	}
	\quad
	\subfloat[][BTI aging when de-stressing a neuron circuit once every 100ms. \label{fig:nbti_demo_2}]{
		\includegraphics[width=1.0\columnwidth]{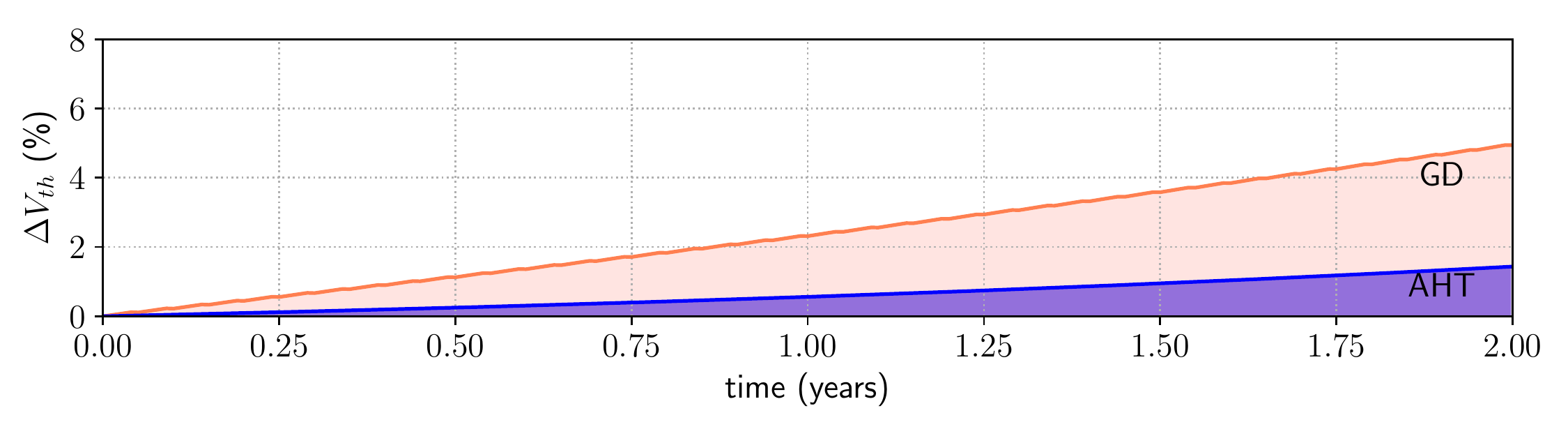}
	}
	\caption{Shift in threshold voltage of a NMOS transistor in a neuron with a constant firing rate of 50Hz and the neuron circuit de-stressed a) once every second (long-term) and b) once every 100ms (short-term).}
	\label{fig:nbti_aging_demo}
\end{figure}

Figure~\ref{fig:nbti_demo_1} shows the shift in threshold voltage of a NMOS transistor in a neuron with a constant firing rate of 50Hz and the neuron circuit de-stressed once every second (see Section \ref{sec:evaluation} for the simulation setup). Figure~\ref{fig:nbti_demo_2} shows the results using the same setup, but with the neuron circuit de-stressed once every 100ms. 
As this figure clearly shows,
with longer de-stress interval (e.g., once every second), the transistor aging becomes irreversible. 
Therefore, the shift in threshold voltage of the transistor is higher than the case with shorter de-stress interval (e.g., once every 100ms).

\section{Run-time Manager for Neuromorphic Computing (\tech{})}\label{sec:rtm}
\subsection{A Motivating Example Showing the Need for Run-time Reliability Management}\label{sec:motivating_example}
\begin{figure}[h!]
 	\centering
    \vspace{-6pt}
 	\centerline{\includegraphics[width=0.99\columnwidth]{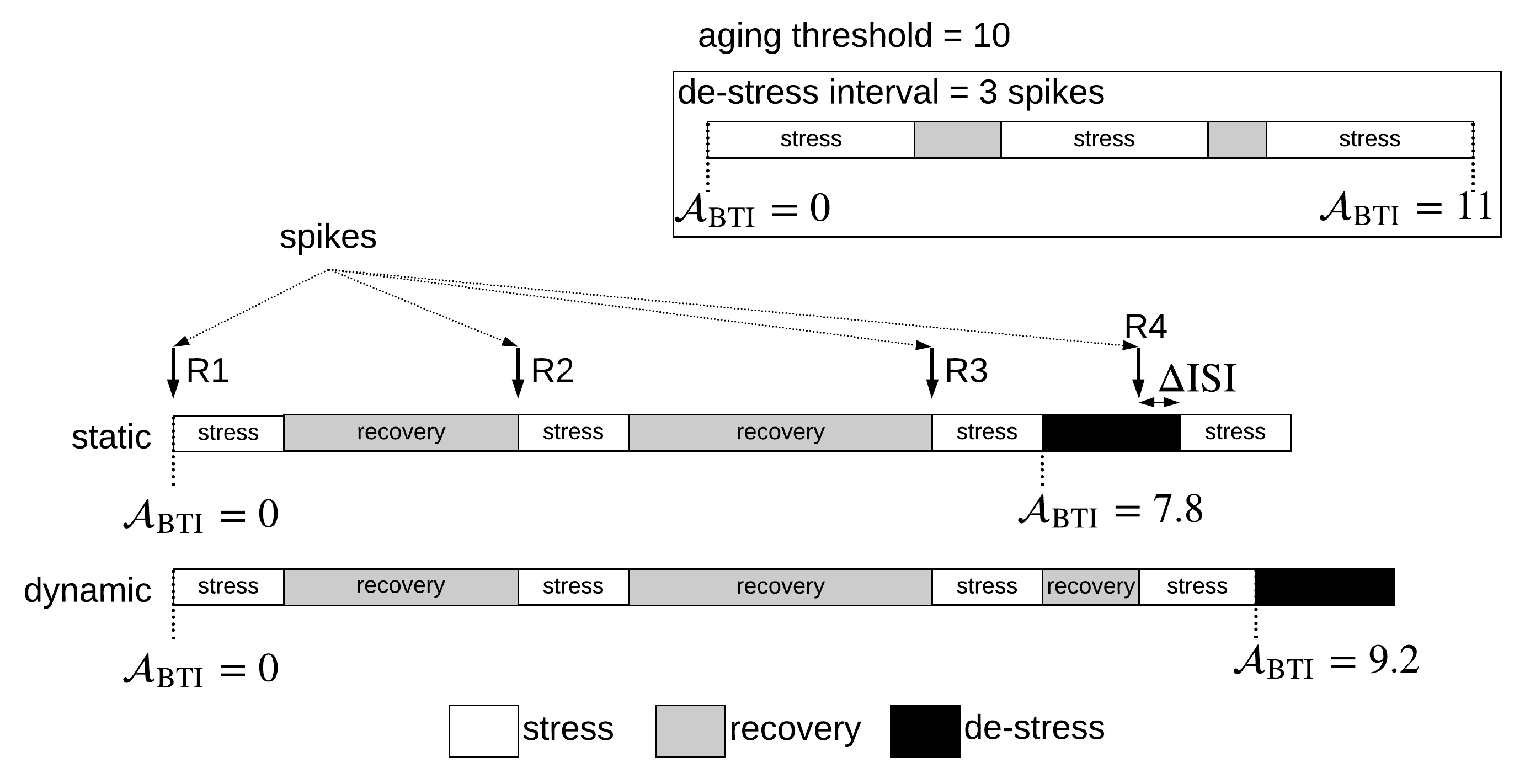}}
 	\vspace{-10pt}
 	\caption{Comparing static vs. dynamic reliability management policy.} 
 	\label{fig:motivation_static_dynamic}
\end{figure}

\rev{
Figure~\ref{fig:motivation_static_dynamic} shows an example where four spikes  (R1, R2, R3, \& R4) are generated from a neuron. These spikes are generated with some idle time between them (based on its input excitation).
The figure illustrates the hybrid approach DTRO~\cite{frameworkCAL} (see Fig.~\ref{fig:sota}b), where the neuron is de-stressed at run-time after generating 3 spikes. This fixed number is decided statically, considering the neuron's activation in some training example. 
This is illustrated in the top right corner of the figure, where we observe that the BTI aging (\ineq{\mathcal{A}_\text{BTI}}) exceeds the aging threshold of 10 units after generating three spikes based on the idle periods between spikes in the training example. 
}

\rev{
Using this static approach, the de-stress operation is initiated upon generating R3, which delays the generation of R4 due to the non-zero latency of the de-stress operation. This causes a change in the ISI, which may lead to performance loss in SNNs (see Appendix~\ref{sec:performance_loss}).
At the time when the neuron circuit is being de-stressed, the \textit{BTI aging is below the threshold} because a CMOS transistor recovers partially from BTI stress when idle. The length of the idle periods at run-time can be different from those used at design-time when the analysis is performed as shown in this example. Therefore, the static approach will unnecessarily introduce performance penalty in such a situation.
}
\nrev{
Using fixed interval for de-stress (instead of counting the spikes) will also lead to a similar situation because the number of spikes within the de-stress interval still remains unknown at run-time, being dependent on the input excitation.
}

\rev{
Figure~\ref{fig:motivation_static_dynamic} also shows \tech{}, our dynamic reliability management policy, where the de-stress operation of the neuron circuit is initiated by tracking its aging. \tech{} can generate the spike R4 because the aging of the neuron is lower than the aging threshold at the time of generating the spike. This is because \tech{} models both the stress and recovery of circuit aging at run-time. There is no change in ISI. Therefore, \tech{} is better than the static approach both in terms of reliability and performance.
}

\nrev{
This example demonstrates one scenario with sparse neuron activation. One can also imagine a counter scenario where the neuron is activated too frequently.
In this case, the static policy can lead violating the critical threshold because it cannot adapt the de-stress interval at run-time. \tech{} can adjust its de-stress interval at run-time by tracking the aging (both stress and recover). In Section~\ref{sec:isi_distortion} we show only a marginal performance impact for workloads with frequent activation. Therefore, \tech{} is better than the static policy, when it comes to managing workload-specific circuit aging.
}

\subsection{High-level Overview}
\begin{figure}[h!]
	\centering
	\centerline{\includegraphics[width=1.0\columnwidth]{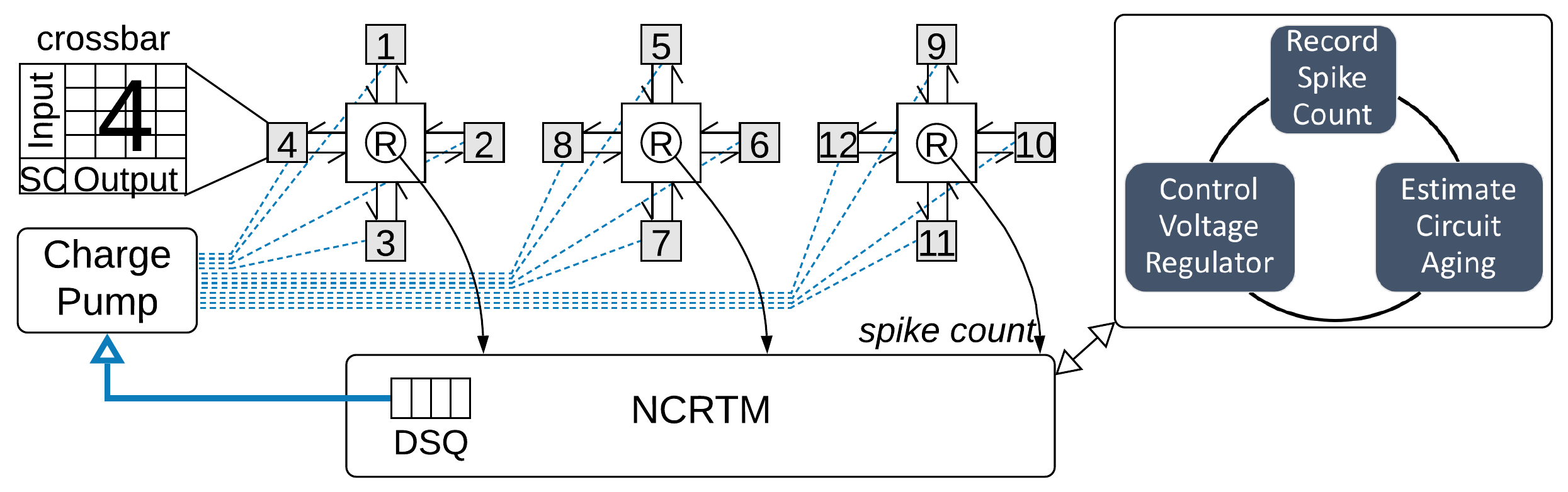}}
	\caption{Overview of \tech{} for tile-based neuromorphic hardware.}
	\label{fig:rtm}
\end{figure}

Based on the three observations in Section~\ref{sec:observations} and the motivating example in Section~\ref{sec:motivating_example}, we introduce \tech{}, a run-time reliability manager for neuromorphic hardware. Figure~\ref{fig:rtm} illustrates \tech{} designed for tile-based neuromorphic hardware. We show the architecture of DYNAP-SE with 12 tiles, numbered \ineq{1,2,\cdots,12} in the figure. These tiles are connected hierarchically, with groups of 4 tiles connected to a local router R. The local routers can be interconnected via global routers (not shown in the figure) to facilitate spike communication between any two tiles. The figure also shows the on-chip charge pump and the voltage delivery network, which supply the reference voltages for the neuron circuits of each tile.

\tech{} is implemented as a controller to mitigating the aging of neuron circuits in each tile. To do so, \tech{} estimates maximum aging of the neurons in each tile by recording the number of spikes within a time window (see our aging formulation in Section~\ref{sec:aging}). If the aging of a tile exceeds a threshold (\ineq{th\_a}),\footnote{The aging threshold \ineq{th_a} is a user-defined parameter used to achieve a given reliability target.} \tech{} schedules the de-stress of the tile by making an entry in the de-stress queue (DSQ). However, the tile may not be de-stressed immediately. \tech{} de-stresses a tile opportunistically by estimating the change in ISI (called ISI distortion) that may result due to offlining the tile (see our ISI formulation in Section~\ref{sec:isi}).
During de-stresses, a very low voltage is applied to all the neurons in a tile for a time duration \ineq{tDSC} (discharge cycle time). This allows the transistors in the neurons to reverse the threshold voltage drift \ineq{\Delta V_\text{th}}. The recovery time \ineq{tDSC} is modeled using the framework presented in~\cite{6531944}.

Fundamental to the aging and ISI computation in \tech{} is a technique to estimate the number of spikes for each neuron. 
\nrev{
The spike counter (SC) in each tile can facilitate counting the spikes in a time interval. However, not all neuromorphic hardware is equipped with SC.
}
Therefore, we present an alternative software-based 
technique for implementing spike counting.

\subsubsection{Spike Counting in Software}
\begin{figure}[h]
	\centering
	\centerline{\includegraphics[width=0.69\columnwidth]{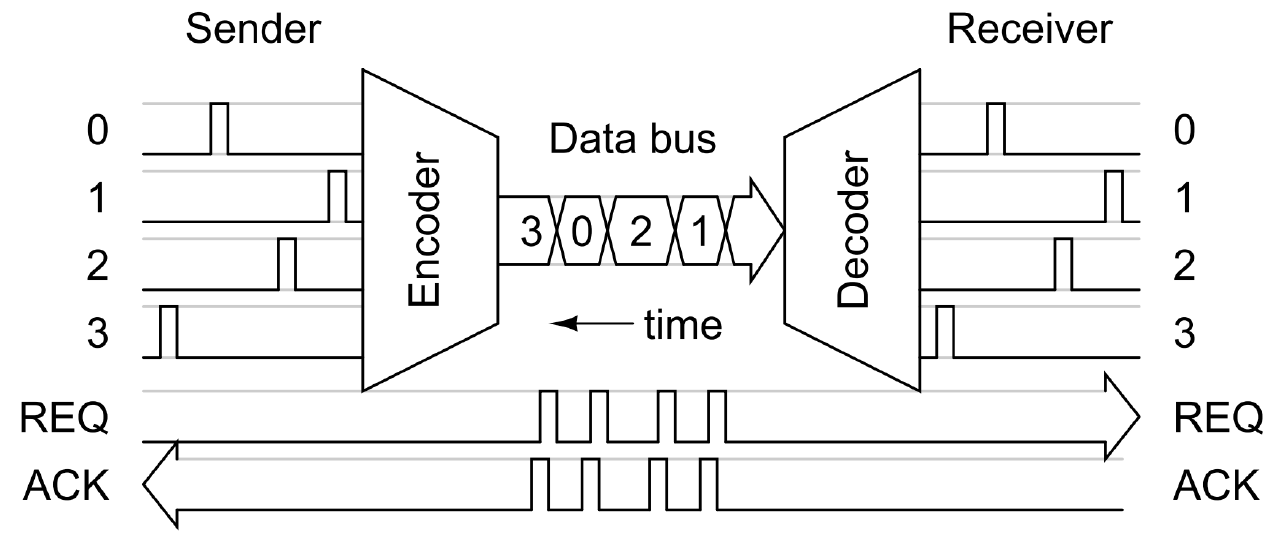}}
	\caption{An example AER protocol.}
	\label{fig:aer}
\end{figure}

To understand spike counting in software, we describe the spike communication mechanism in neuromorphic hardware. 
Spikes from the post-synaptic neurons in a tile are converted into an address-event representation (AER) and broadcasted on the interconnect via the AER encoder.
Figure~\ref{fig:aer} shows an example explaining the principles behind AER. Here, four neurons in a tile spike at time 3, 0, 1 and 2 time units, respectively. The encoder encodes these four spikes in order to be communicated on the interconnect. 

We propose to count the spikes from each neuron by snooping on the interconnect. We implement counters, one for each neuron of the hardware. When a de-stress operation is initiated for a tile, all the counters for the neurons in the tile are reset to start counting the spikes for the next interval. \mr{The total storage overhead needed for implementing software-based spike counting for the 12-tile architecture of Figure~\ref{fig:rtm} is \ineq{12*128*16} bits = 24Kb, with 128 post-synaptic neurons per tile.} However, continuous snooping on the bus can introduce performance overhead. 
In the future, we plan to extend the interconnect routers to facilitate recording the spike packets. This will allow \tech{} to poll these readings periodically.
With this necessary background, we now introduce our model for estimating aging (Section~\ref{sec:aging}) and ISI (Section~\ref{sec:isi}).

\subsection{Aging Computation}\label{sec:aging}
Equation~\ref{eq:MTTF_NBTI} equates the MTTF of a CMOS transistor for a given overdrive voltage.
BTI failures can also be modeled using the Weibull distribution with a scale parameter \ineq{\alpha} and a slope parameter \ineq{\beta}. Reliability at time \ineq{t} can be written as
\begin{equation}
\label{eq:eq1}
\scriptsize R(t) = e^{-\left(\frac{t}{\alpha(V)}\right)^\beta},
\end{equation}
with the corresponding MTTF computed as 
\begin{equation}
\label{eq:2}
\scriptsize MTTF = \int_{0}^{\infty}R(t)dt = \alpha(V)\Gamma\left(1+ \frac{1}{\beta}\right),
\end{equation}
where \ineq{\Gamma} is the Gamma function.
Using the expressions for MTTF from Equations~\ref{eq:MTTF_NBTI} and \ref{eq:eq1}, and rearranging, we obtain the expression for the scale parameter \ineq{\alpha} as
\begin{equation}
\label{eq:eq3}
\scriptsize \alpha(V) = \frac{\frac{A}{V^\gamma}e^{\frac{E_a}{KT}}}{\Gamma\left(1+\frac{1}{\beta}\right)}.
\end{equation}
The aging (\ineq{\mathcal{A}}), i.e., the degradation of the CMOS transistor can be expressed as
\begin{equation}
\label{eq:aging}
\scriptsize \mathcal{A} = \sum_{i=1}^{n}\frac{\Delta t_i}{\alpha(V_i)}, \text{ such that } R(t_s) = e^{-(\mathcal{A})^{\beta}},
\end{equation}
where the scaling factor \ineq{\alpha(V_i)} can be calculated using Eq. \ref{eq:eq3}.

\mr{
We note that a neuron suffers aging when generating a spike. Each spike is of fixed voltage \ineq{V_{spk}} (see Figure~\ref{fig:neuron}) and a fixed time duration to the order of few ms. Therefore, both \ineq{V_i} and \ineq{t_i} in Equation~\ref{eq:aging} are constant. This allows us to express the aging formulation as
\begin{equation}
\label{eq:aging_final}
\scriptsize \mathcal{A} = n\cdot\frac{\Delta t}{\alpha(V)},
\end{equation}
where \ineq{n} is the number of spikes generated by the neuron, \ineq{\Delta t} is the fixed spike duration, and \ineq{V} is the fixed spike voltage. Equation~\ref{eq:aging_final} allows us to represent the aging in terms of the number of spikes generated by a neuron and the unit aging parameter \ineq{\frac{\Delta t}{\alpha(V)}}, which represents the aging per spike. This simplified formulation allows to estimate the aging in each neuron by simply counting the number of spikes it generates.
Hence the performance overhead can be kept negligible.
}

\subsection{ISI Computation}\label{sec:isi}
To define ISI, we consider a tile consisting of \ineq{N} post-synaptic neurons and a finite interval of time \ineq{[0,T]} for which the tile is active without undergoing a de-stress operation. The post-synaptic neurons generate \ineq{K} spikes in this interval, which are organized based on their generation time and the source neuron as
\begin{equation}
    \label{eq:spike_time}
    \footnotesize  \{t_1^1,t_2^1,\cdots,t_{k_1}^1\},\{t_1^2,t_2^2,\cdots,t_{k_2}^2\},\cdots,\{t_1^N,t_2^N,\cdots,t_{k_N}^N\},
\end{equation}
where \ineq{t_i^n} is the time of the \ineq{i^\text{th}} spike generated by the \ineq{n^\text{th}} neuron and \ineq{K = \sum_{i=1}^N k_i}. The instantaneous ISI of the spike train from the \ineq{n^\text{th}} neuron is \cite{grun2010analysis}
\begin{equation}
    \label{eq:isi}
    \footnotesize ISI_\text{inst}^n(i) = t_i^n - t_{i-1}^n
\end{equation}

To estimate the impact of de-stress operation on ISI, we compute two statistics for each neuron -- the instantaneous ISI (Equation~\ref{eq:isi}) and the average ISI, which is computed as the average of all ISIs for a neuron. Using \ineq{tDSC} as the time to de-stress a tile, the change in instantaneous and average ISI of the \ineq{n^\text{th}} neuron are 
\begin{equation}
    \label{eq:isi_final}
    \footnotesize \Delta ISI_\text{inst}^n(i) = ISI_\text{inst}^n(i) + tDSC \text{ and } \Delta ISI_\text{avg}^n = tDSC / k_N
\end{equation}






\section{Evaluation Methodology}\label{sec:evaluation}
\begin{figure}[h]
	\centering
	\vspace{-5pt}
	\centerline{\includegraphics[width=0.79\columnwidth]{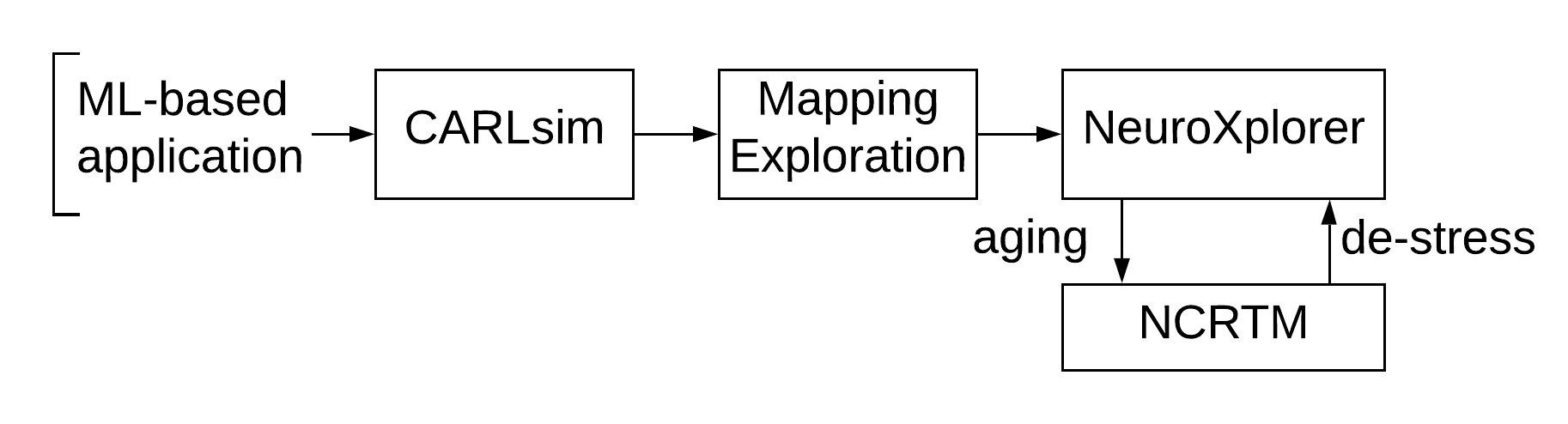}}
	\vspace{-10pt}
	\caption{Our simulation framework.}
	\label{fig:simulation_framework}
\end{figure}

\rev{
Figure~\ref{fig:simulation_framework} illustrates our simulation framework. An SNN-based application is simulated using \texttt{CARLsim}~\cite{carlsim}, a GPU accelerated SNN simulator used to train and test SNN models. CARLsim reports spike times for every synapse in the SNN. The spike times are used to perform mapping explorations optimizing some objective, such as performance (PyCARL~\cite{pycarl}) and reliability (RENEU~\cite{reneu}). We use \texttt{NeuroXplorer}~\cite{neuroxplorer}, a cycle-accurate simulator of neuromorphic hardware such as DYNAP-SE~\cite{moradiDynap2017}. The neuron and synapse mapping obtained using the mapping exploration framework is applied to \texttt{NeuroXplorer} to perform cycle-accurate simulation of the application on the hardware model, using current data. \tech{} is implemented inside \texttt{NeuroXplorer} to estimate the circuit aging and control it by de-stressing the circuit when the aging exceeds a threshold. The change in spike latency due to the de-stress operation can be precisely modeled in \texttt{NeuroXplorer}, as shown in \cite{neuroxplorer}.
}

All simulations are conducted on a system with 8 CPUs, 32GB RAM, and NVIDIA Tesla GPU, running Ubuntu 16.04. 

\subsection{Evaluated Applications}
We evaluated 10 machine learning applications that are representative of three most commonly used neural network classes --- convolutional neural network (CNN), multi-layer perceptron (MLP), and recurrent neural network (RNN).
These applications are 
1) LeNet~\cite{lenet} based handwritten digit recognition with \ineq{28 \times 28} images of handwritten digits from the MNIST dataset \cite{deng2012mnist};
2) AlexNet~\cite{alexnet} for Imagenet classification~\cite{deng2009imagenet};
3) VGG16~\cite{vgg16}, also for Imagenet classification~\cite{deng2009imagenet};
4) ECG-based heart-beat classification (HeartClass)~\cite{das2018heartbeat,HeartClassJolpe} using electrocardiogram (ECG) data from the Physionet database~\cite{moody2001physionet};
5) {multi-layer perceptron (MLP)-based handwritten digit recognition} (MLP-MNIST)~\cite{Diehl2015} using the MNIST database;
6) {image smoothing} (ImgSmooth)~\cite{carlsim} on $64 \times 64$ images; 
7) {edge detection} (EdgeDet)~\cite{carlsim} on $64 \times 64$ images using difference-of-Gaussian; 
8) {heart-rate estimation} (HeartEstm)~\cite{HeartEstmNN} using ECG data;
9) gender classification using speech data (SpeechRecog)~\cite{dong2018unsupervised}; and
10) RNN-based predictive visual pursuit (VisualPursuit)~\cite{Kashyap2018}. 
The former 7 are supervised applications, while the latter 3 are unsupervised applications.
Table~\ref{tab:apps} summarizes the topology, the number of neurons and synapses of these applications, and their baseline accuracy on DYNAP-SE using PyCARL~\cite{pycarl}.\footnote{\mr{The CNN models LeNet, AlexNet, and VGG16 are converted to spiking domain using our previously proposed converter~\cite{HeartClassJolpe}. For the inference performance of the original model, readers are referred to~\cite{reddi2020mlperf}.}}

\begin{table}[h!]
	\renewcommand{\arraystretch}{1.2}
	\setlength{\tabcolsep}{2pt}
	\caption{Applications used to evaluate our approach \tech{}.}
	\label{tab:apps}
	\vspace{-10pt}
	\centering
	\begin{threeparttable}
	{\fontsize{8}{12}\selectfont
		\begin{tabular}{cc|ccl|c}
			\hline
			\textbf{Class} & \textbf{Applications} & \textbf{Synapses} & \textbf{Neurons} & \textbf{Topology} & \textbf{Accuracy}\\
			\hline
			\multirow{4}{*}{CNN} & LeNet \cite{lenet} & 159,553 & 5,576 & CNN & 94.08\%\\
			& AlexNet \cite{alexnet} & 1,029,286 & 650,000 & CNN & 71.7\%\\
			& VGG16 \cite{vgg16} & 2,136,560 & 18,472 & CNN & 91.62 \%\\
			& HeartClass \cite{HeartClassJolpe} & 2,396,521 & 24,732 & CNN & 85.12\%\\
			\hline
			\multirow{3}{*}{MLP} & MLP-MNIST \cite{Diehl2015} & 79,400 & 984 & FeedForward (784, 100, 10) & 95.5\%\\
			& EdgeDet \cite{carlsim} & 272,628 &  1,372 & FeedForward (4096, 1024, 1024, 1024) & 100\%\\
			& ImgSmooth \cite{carlsim} & 136,314 & 980 & FeedForward (4096, 1024) & 100\%\\
			\hline
 			\multirow{3}{*}{RNN} & HeartEstm \cite{HeartEstmNN} & 636,578 & 6,952 & Recurrent Reservoir & 99.2\%\\
 			& SpeechRecog \cite{dong2018unsupervised} & 39,056 & 683 & Recurrent Reservoir & 96.8\%\\
 			& VisualPursuit \cite{Kashyap2018} & 3,25,710 & 5,717 & Recurrent Reservoir & 89.0\%\\
			\hline
	\end{tabular}}
	\end{threeparttable}
	\vspace{-10pt}
\end{table}

\subsection{Hardware Models}
\minorrev{
In our cycle-accurate simulator, 
we model the architecture of the DYNAP-SE neuromorphic hardware~\cite{moradiDynap2017} with the following configurations.
}

\begin{itemize}
    \item A tiled array, with each tile accommodating 128 input and 128 output neurons. There are 65,536 crosspoints in each tile.
    \item Spikes are digitized and communicated between cores through a mesh routing network using the Address Event Representation (AER) protocol.
\end{itemize}

\minorrev{
The DYNAP-SE platform uses static random access memory (SRAM) to implement the synaptic cells in each crossbar. However, in our simulator, we use Phase-Change Memory (PCM) as the synaptic element. Table~\ref{tab:hw_parameters} reports the major hardware parameters.
\begin{table}[h!]
    \caption{\minorrev{Major simulation parameters extracted from \cite{moradiDynap2017} and extrapolated for PCM technology.}}
	\label{tab:hw_parameters}
	\centering
	{\fontsize{6}{10}\selectfont
		\begin{tabular}{lp{5cm}}
			\hline
			Neuron technology & 65nm CMOS\\
			\hline
			Synapse technology & PCM\\
			\hline
			Supply voltage & 1.0V\\
			\hline
			Energy per spike & 50pJ at 30Hz spike frequency\\
			\hline
			Energy per routing & 147pJ\\
			\hline
			Switch bandwidth & 1.8G. Events/s\\
			\hline
	\end{tabular}}
\end{table}
}

\minorrev{
In the future, we will demonstrate \tech{} on a real NVM-based silicon neuromorphic system.
}


\subsection{Evaluated State-of-the-art Techniques}
We evaluate the following three approaches.
\begin{itemize}
    \item \textbf{PyCARL~\cite{pycarl}:} This is a performance-oriented approach to map SNN-based applications to neuromorphic hardware. This approach first generates clusters of neurons and synapses, where each cluster can fit on to the resources of a tile in the hardware. Then it uses an optimization algorithm to place these clusters to the hardware, maximizing performance of the machine learning application on the hardware. CMOS circuits are not de-stressed at run-time.
    \item \textbf{RENEU~\cite{reneu}:} This is a reliability-oriented approach to map SNN-based applications to neuromorphic hardware. This approach also generates clusters of neurons and synapses from an application, but maps the clusters to the hardware minimizing the maximum aging while considering only the training data. CMOS circuits are not de-stressed at run-time.
    \item \textbf{DTRO~\cite{frameworkCAL}:} This is a reliability-oriented approach where neuron and synapse circuits of the neuromorphic hardware are de-stressed at run-time at fixed interval. This interval is decided based on analysis performed at design-time using training data.
\end{itemize}

\subsection{Evaluated Metric}
We evaluate the following metrics.
\begin{itemize}
    \item \textbf{Aging:} This is the maximum circuit aging in DYNAP-SE for each machine learning workload.
    \item \textbf{ISI:} This is average ISI of each machine learning workload.
    \item \minorrev{
    \textbf{Application Performance (accuracy):} The performance, e.g., accuracy is defined in terms of misclassification rate for image-based CNN and MLP applications. For RNN applications that use time-series data, performance is measured in terms of error rate~\cite{HeartEstmNN}.
    }
    \item \textbf{Aging per unit ISI distortion:} This is an unified metric reporting the aging per unit ISI distortion for each workload, defined as
    \begin{equation}
        \label{eq:dist_aging}
        \footnotesize \text{aging per unit ISI distortion} =  \mathcal{A} / \Delta ISI
    \end{equation}
\end{itemize}

\minorrev{
In formulating the optimization objective of Equation~\ref{eq:dist_aging}, \tech{} aims to optimize (i.e., minimize) circuit aging \ineq{\mathcal{A}} for a given constraint on the ISI distortion.
In our earlier works~\cite{psopart,spinemap}, we have shown the dependency of application performance, e.g., accuracy on ISI due to inter-spike interval-based information encoding in SNNs.
Therefore, any distortion in ISI may lead to a reduction in performance~\cite{spinemap,psopart}.
Correspondingly,
the above optimization problem essentially reduces to minimizing the aging \ineq{\mathcal{A}} for a given constraint on SNN accuracy.
}

\subsection{Aging Parameters}
To compute aging, the slope parameter of Weibull distribution is set to \ineq{\beta = 2}, and the operating temperature is set to \ineq{300K}. Other fitting parameters are adjusted to achieve an MTTF of {2 years} in the baseline system (PyCARL), corresponding to a threshold voltage shift of 10\%. This is what is typically accepted by technologists as the maximum allowed degradation before timing errors begin to appear.

\section{Results and Discussion}\label{sec:results}
\subsection{Summary of Results}
\label{sec:summary}
Table \ref{tab:compare_sota} summarizes the key results.

\begin{table}[h!]
	\renewcommand{\arraystretch}{1.2}
	\setlength{\tabcolsep}{2pt}
	\caption{Summary of results.}
	\label{tab:compare_sota}
	\vspace{-10pt}
	\centering
	{\fontsize{8}{12}\selectfont
		\begin{tabular}{|l|c|c|c|c|}
			\hline
			\multirow{2}{*}{\textbf{\tech{}}} & \textbf{Aging} & $\mathbf{\Delta V_\text{th}}$ & $\mathbf{\Delta}$ \textbf{ISI} & \textbf{Accuracy} \\
			& \textbf{(Sec. \ref{sec:aging_results})} & \textbf{(Sec. \ref{sec:vth_shift})} & \textbf{(Sec. \ref{sec:isi_distortion})} & \textbf{(Sec. \ref{sec:accuracy})}\\
			\hline
vs. PyCARL~\cite{pycarl} & 74\%$\downarrow$ & 52.0\%$\downarrow$ & 12\%$\uparrow$ & 4.52\%$\downarrow$ \\
vs. RENEU~\cite{reneu} & 73\%$\downarrow$ & 50.7\%$\downarrow$ & 11\%$\uparrow$ & 4.52\%$\downarrow$ \\
vs. DTRO~\cite{frameworkCAL} & 60\%$\downarrow$ & 31.4\%$\downarrow$ & 2\%$\uparrow$ & 0.76\%$\downarrow$ \\
			\hline
	\end{tabular}}
\end{table}

\subsection{Circuit Aging}\label{sec:aging_results}
\begin{figure}[h!]
	\centering
	\vspace{-5pt}
	\centerline{\includegraphics[width=0.99\columnwidth]{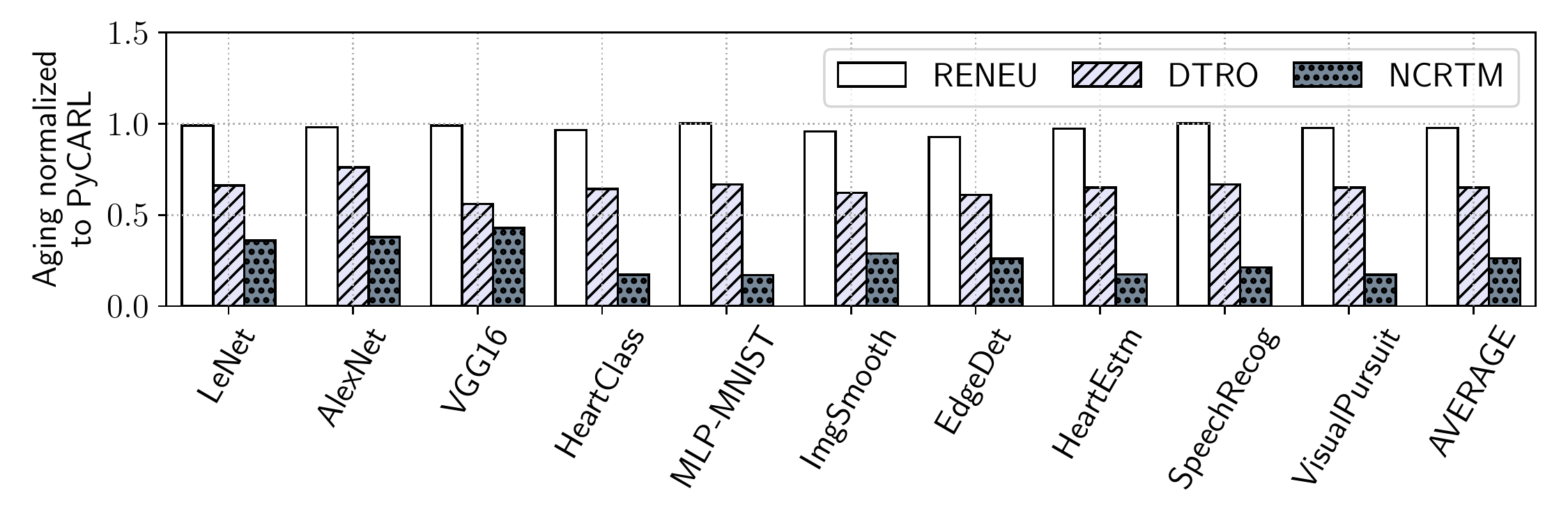}}
	\vspace{-5pt}
	\caption{Aging at 300K normalized to PyCARL.}
	\label{fig:aging}
\end{figure}

Figure~\ref{fig:aging} plots the aging of the neuron and synapse circuits in DYNAP-SE during the execution of the machine learning applications for each evaluated approach, normalized to PyCARL. We make the following three key observations.

First, the aging due to RENEU is lower than PyCARL by an average of 2.5\%. This improvement is due to the aging-aware neuron and synapse mapping policy of RENEU, which balances the aging of all tiles in the hardware. PyCARL, which balances the utilization of the tiles in the hardware, has higher aging. However, both PyCARL and RENEU are design-time based policies, i.e., they do not make any run-time decisions. 
Second, DTRO is a hybrid approach, which uses the neuron and synapse mapping of RENEU. Compared to PyCARL and RENEU, DTRO de-stresses all circuits in the hardware periodically at run-time. The de-stress interval is determined at design-time by analyzing the training data. The aging of DTRO is therefore lower than both PyCARL (average 35\% lower) and RENEU (average 33.5\% lower). 
Third, \tech{}, which is a run-time approach, has the lowest aging of all. The average aging of \tech{} is 74\% lower than PyCARL, 73\% lower than RENEU, and 60\% lower than DTRO. The improvement of \tech{} is due to the precise tracking of aging at run-time using current data, to achieve a target MTTF.

\subsection{Threshold Voltage Shift}\label{sec:vth_shift}
\begin{figure}[h!]
	\centering
	\vspace{-5pt}
	\centerline{\includegraphics[width=0.99\columnwidth]{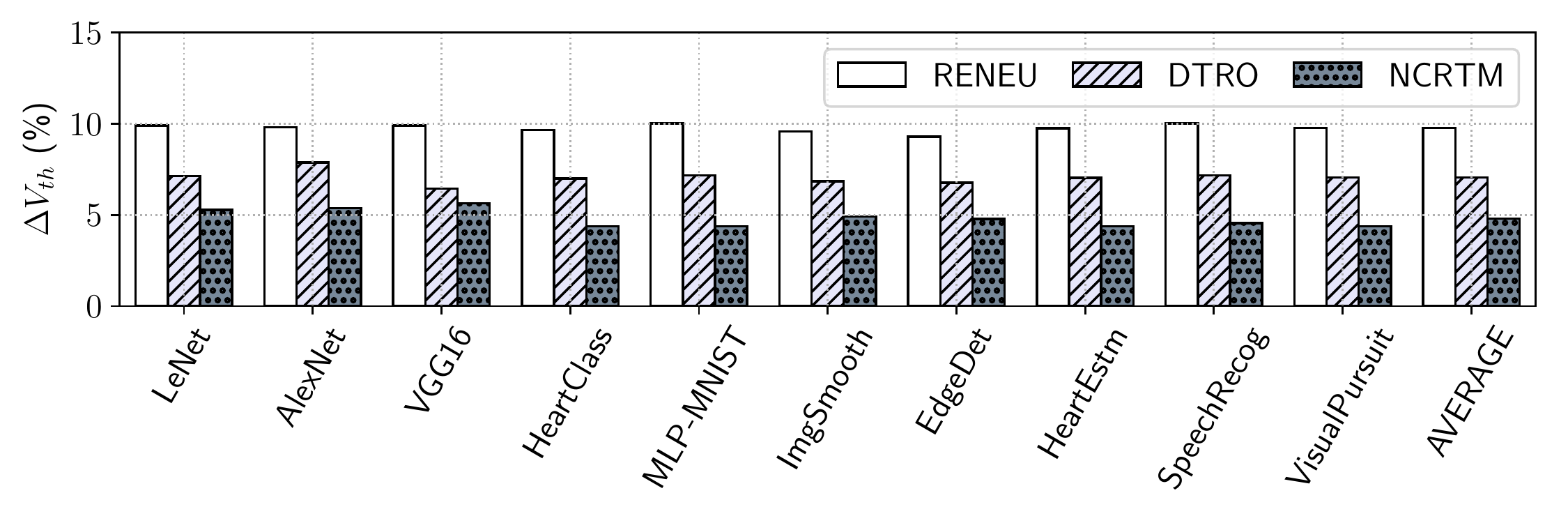}}
	\vspace{-5pt}
	\caption{Threshold voltage shift after 2 years of continuous operation.}
	\label{fig:vth}
\end{figure}

Circuit aging manifests as shift in threshold voltage. Figure~\ref{fig:vth} plots the shift in threshold voltage (\ineq{\Delta V_\text{th}}) in DYNAP-SE after executing each machine learning application continuously till the end of its lifetime of 2 years.
We normalize the results so that the threshold voltage shift due to PyCARL is 10\%.
We make the following key observations. 

We observe that, compared to PyCARL, the average threshold voltage shift when using RENEU is 9.75\%, DTRO is 7.0\%, and \tech{} is only 4.8\%. The threshold voltage shift is the lowest in \tech{} because the aging of \tech{} is lowest of all the approaches, which we reported in Section~\ref{sec:aging_results}. 
Increase in threshold voltage results in the reduction in drive current, which in turn results in temporal performance degradation of neuron and synapse circuits in the neuromorphic hardware.

\subsection{Change in ISI}\label{sec:isi_distortion}
\begin{figure}[h!]
	\centering
	\vspace{-5pt}
	\centerline{\includegraphics[width=0.99\columnwidth]{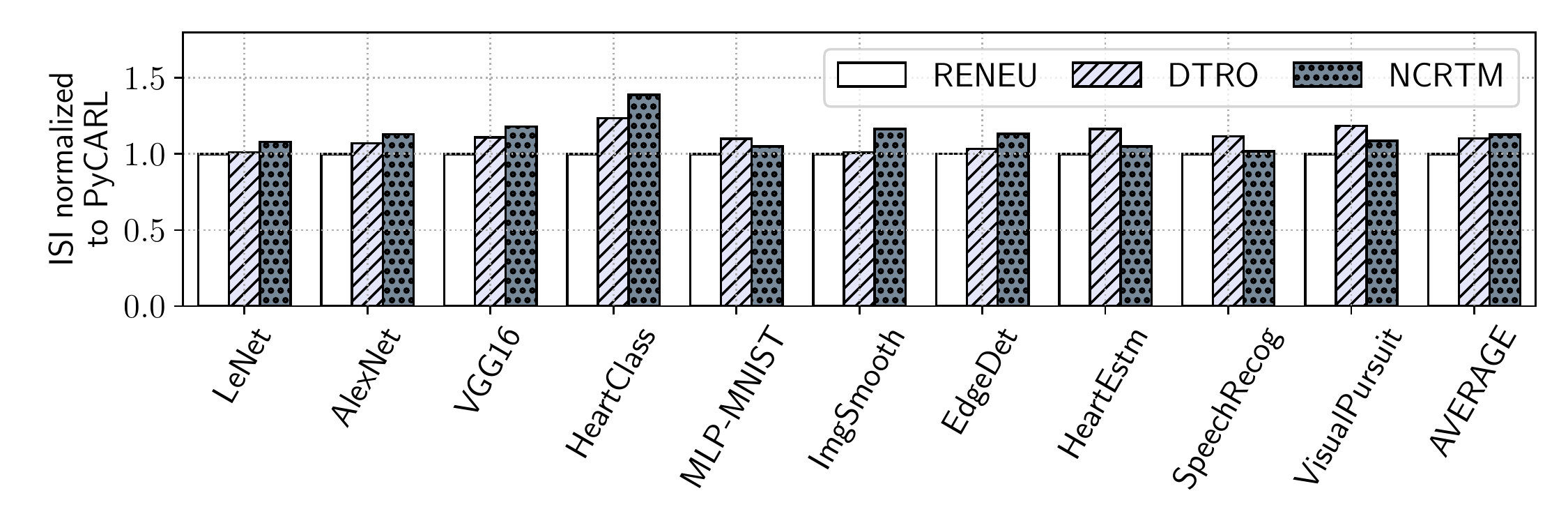}}
	\vspace{-5pt}
	\caption{ISI at 300K normalized to PyCARL.}
	\label{fig:isi_results}
\end{figure}

\nrev{
Figure~\ref{fig:isi_results} plots the ISI of the machine learning workloads on DYNAP-SE for each evaluated approach, normalized to PyCARL. We make the following five key observations.
}

First, the ISI obtained with RENEU is similar to PyCARL. 
This is because RENEU generates a mapping of the workload to the hardware, which improves reliability without significantly hurting the performance. Since no run-time decisions are made in both these approaches, their performance at run-time are therefore similar.
Second, the ISI obtained with DTRO is higher than RENEU by an average of 10\%. This increase is because DTRO make run-time decision of de-stressing the neuron and synapse circuits periodically to control their aging. This leads to increase in latency, which increases the average ISI (see Equation~\ref{eq:isi_final}). The advantage in this case is lower aging, which we analyzed in Section~\ref{sec:aging_results}, leading to a lower drift of the threshold voltage (see Section~\ref{sec:vth_shift}).
Third, 
the ISI using \tech{} is higher than RENEU by an average of 12\%. 
This increase is due to the run-time de-stresses in \tech{} (similar to DTRO), which introduces latency, impacting the ISI.
The ISI using \tech{} is only 2\% higher than DTRO. This increase is because \tech{} never allows the aging to reach critical levels and therefore, schedules more de-stresses by precisely tracking it at run-time. However, due to \tech{}'s policy to schedule the de-stresses by tracking their latency impact on ISI, \tech{} ensures only marginal change in ISI. ISI change may lead to accuracy impact, which is discussed in Section~\ref{sec:accuracy}.
Fourth, the ISI of \tech{} is lower than DTRO for MLP-MNIST. This is because for this application, the circuit aging is generally lower due to the sparsity of spike generation in the workload. So, the BTI stress is recovered in the idle period. DTRO cannot track this recovery and therefore, applies a conservative control, unnecessarily constraining the performance.
Finally, for the three unsupervised applications (HeartEstm, SpeechRecog, and VisualPursuit), the ISI using \tech{} is on average 10\% lower than DTRO. This is because in the absence of training data for these applications, DTRO applies a conservative policy to de-stress neuron and synapse circuits frequently to prevent their aging from reaching a critical value. \tech{}, on the other hand, tracks the aging at run-time based on the data that these models encounter and de-stress the circuits, only when needed.

\nrev{
\textit{We \textbf{conclude} that for machine learning workloads with sparse activation, \tech{} is significantly better than design-time based approaches both in terms of reliability and performance. For dense activation, \tech{} improves reliability significantly compared to these approaches, with marginal impact on performance. Furthermore, \tech{} outperforms any design-time based policy, when the availability of training data is limited.}
}

\subsection{Application Accuracy}\label{sec:accuracy}
Change in ISI manifests as loss in accucay of a machine learning workload.
Table~\ref{tab:accuracy} reports the accuracy of each machine learning workload on DYNAP-SE using the evaluated approaches. \mr{We make the following three key observations. First, the accuracy of RENEU and PyCARL are the same. This is because RENEU maps neurons and synapses of an SNN workload to the hardware resources statically to minimize the aging. It does so, ensuring that the spike communication latency on the interconnect does not induce any change in ISI compared to that obtained using PyCARL. Second, the accuracy of \tech{} is on average 4.52\% lower than PyCARL and RENEU, and 0.76\% lower than DTRO. This reduction in accuracy is a direct result of the change in ISI, which we analyzed in Section~\ref{sec:isi_distortion}. Third, although \tech{} results in 4.8\% lower accuracy than Baseline for AlexNet (71.7\% Baseline accuracy compared to 68.2\% accuracy using \tech{}), it reduces circuit aging by 62\% compared to Baseline (See Sec.~\ref{sec:aging_results}).
}

\begin{table}[h!]
	\renewcommand{\arraystretch}{1.2}
	\setlength{\tabcolsep}{2pt}
	\caption{Application Accuracy.}
	\label{tab:accuracy}
	\vspace{-10pt}
	\centering
	{\fontsize{8}{12}\selectfont
		\begin{tabular}{|l|c|c|c|l|c|c|c|}
			\hline
			\multirow{2}{*}{\textbf{Application}} & \multicolumn{3}{|c|}{\textbf{Accuracy}} & \multirow{2}{*}{\textbf{Application}} & \multicolumn{3}{|c|}{\textbf{Accuracy}}\\
			\cline{2-4}\cline{6-8}
			 & \textbf{RENEU/PyCARL} & \textbf{DTRO} & \textbf{\tech{}} &  & \textbf{RENEU/PyCARL} & \textbf{DTRO} & \textbf{\tech{}}\\
			\hline
			LeNet & 94.08\% & 91.7\% & 90.6\% & AlexNet & 71.7\% & 69.3\% & 68.2\% \\
			VGG16 & 91.62\% & 90.2\% & 90.1\% & HeartClass & 85.12\% & 80.7\% & 80.1\% \\
			MLP-MNIST & 95.5\% & 90.1\% & 90.1\% & ImgSmooth & 100\% & 99\% & 99\% \\
			EdgeDet & 100\% & 96\% & 95\% & HeartEstm & 99.1\% & 92.6\% & 90.0\% \\
			SpeechRecog & 96.8\% & 94.0\% & 94.1\% & VisualPursuit & 89\% & 81.5\% & 81.5\% \\
			\hline
	\end{tabular}}
\end{table}
\vspace{-10pt}


\subsection{Platform Exploration}
\begin{figure}[h!]
	\centering
	\vspace{-5pt}
	\centerline{\includegraphics[width=0.99\columnwidth]{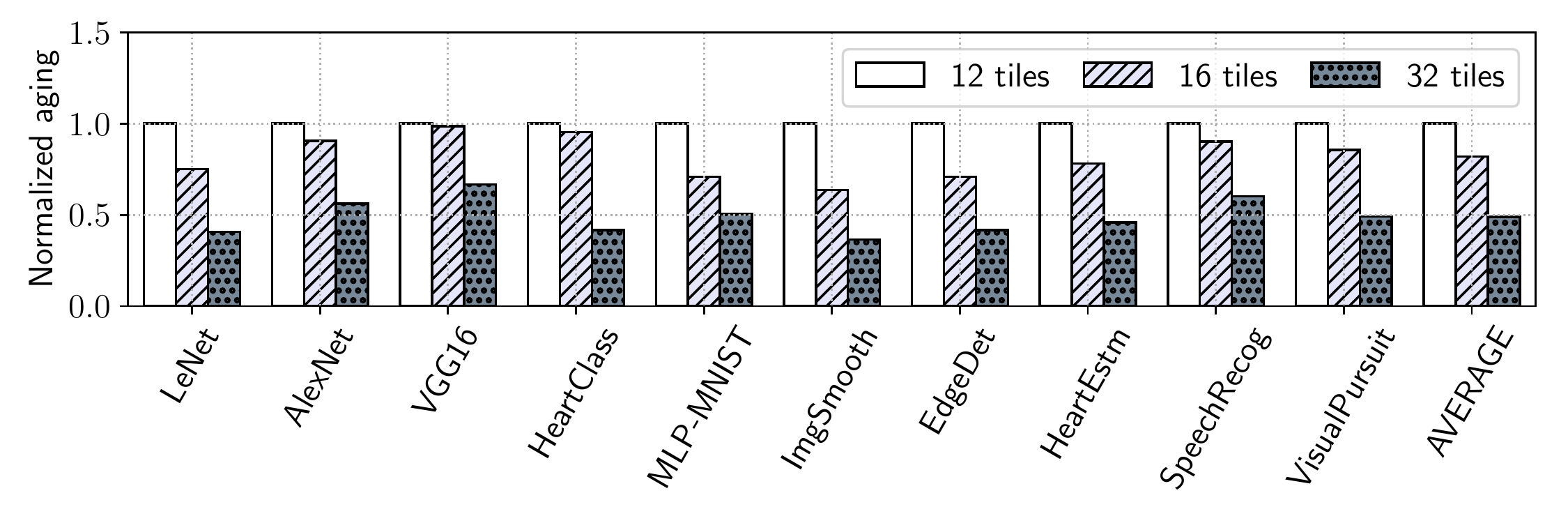}}
	\vspace{-5pt}
	\caption{Aging of DYNAP-SE with 16 and 32 tiles normalized to the aging with 12 tiles.}
	\vspace{-10pt}
	\label{fig:platform}
\end{figure}

Figure~\ref{fig:platform} illustrates the reliability impact of increasing the number of tiles in a neuromorphic hardware. The figure plots the aging results of \tech{} on DYNAP-SE with 16 and 32 tiles, normalized to the aging on DYNAP-SE with the baseline configuration of 12 tiles.
We observe that the average aging with 16 and 32 tiles is 18\% and 51\% lower than the aging with baseline configuration of 12 tiles, respectively. 
Circuit aging is lower with more number of tiles. This is because with more tiles in the hardware, fewer neurons and synapses are mapped to each tile. Therefore, each tile of the hardware generates fewer spikes, which lowers the aging of its neurons and synapse circuits.

\subsection{Temperature Dependency of Reliability}
\begin{figure}[h!]
	\centering
	\vspace{-5pt}
	\centerline{\includegraphics[width=0.99\columnwidth]{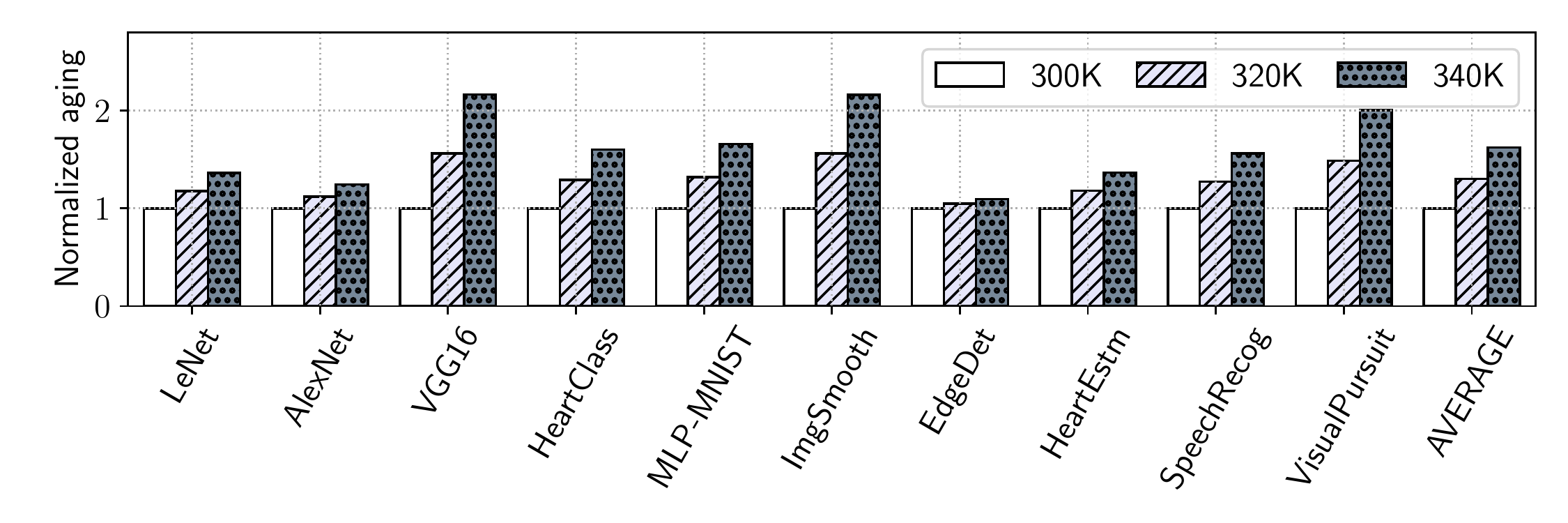}}
	\vspace{-5pt}
	\caption{Aging at 320K and 340K normalized to the aging at 300K.}
	\vspace{-10pt}
	\label{fig:temperature}
\end{figure}

Figure~\ref{fig:temperature} illustrates the temperature dependency of the aging in a neuromorphic hardware. We report the aging results of \tech{} at two elevated temperatures, 320K and 340K, for each of our machine learning applications. Aging results are normalized to \tech{} at 300K.
We observe that aging increases with an increase in temperature. Aging at 320K and 340K is higher than the aging at 300K by an average of 7\% and 30\%, respectively.
This is due to the exponential dependency of circuit aging on temperature (Equation~\ref{eq:aging}). 
We also observe from this equation that aging also depends on the voltage needed to operate the neurons and synapses in the hardware when generating and propagating spikes. Therefore, VGG16, ImgSmooth, and VisualPursuit, which have more spikes, have higher aging at the elevated temperatures than all other applications.
Higher aging leads to higher threshold voltage shift in the transistors.

\subsection{Aging Per Unit ISI Distortion}
\begin{figure}[h!]
	\centering
	\vspace{-5pt}
	\centerline{\includegraphics[width=0.99\columnwidth]{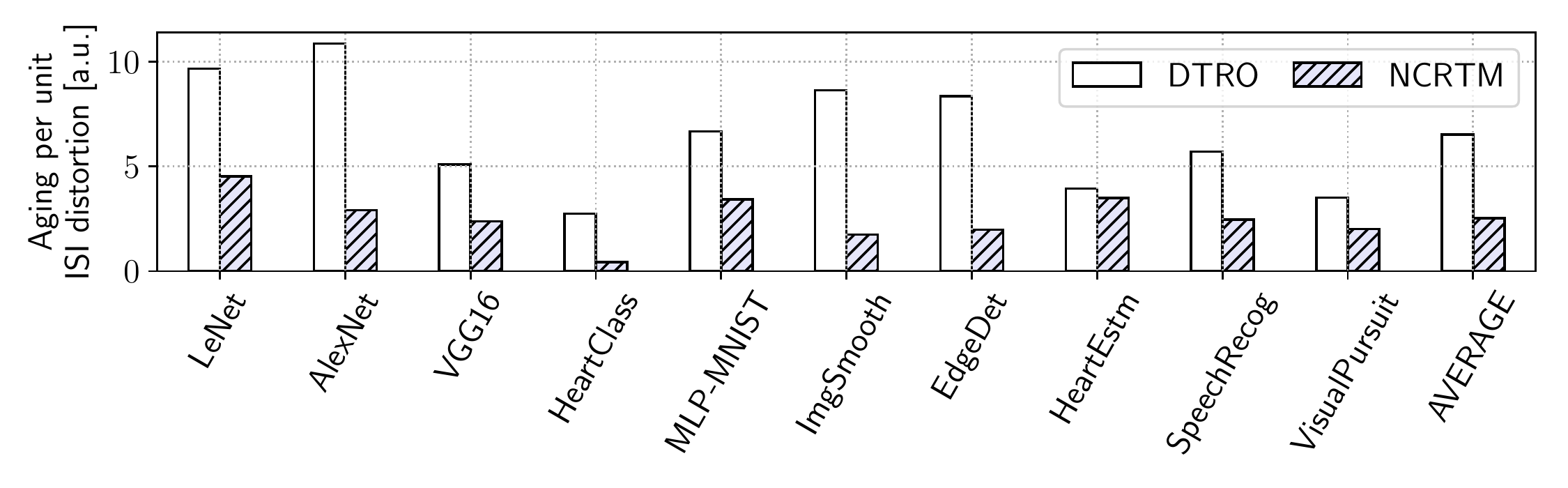}}
	\vspace{-5pt}
	\caption{Aging per unit ISI distortion (lower is better).}
	\label{fig:aging_dist}
\end{figure}

\nrev{
To unify the ISI distortion and aging results in one metric, Figure~\ref{fig:aging_dist} reports the aging per unit distortion of Equation~\ref{eq:dist_aging} for the design-time based RENEU and the run-time based \tech{}.
We observe that \tech{} has an average 58\% lower aging per unit ISI distortion than RENEU. The improvement of \tech{} is what we have analyzed before. This 
result shows that \textit{for the same amount of ISI distortion (i.e., performance impact), \tech{} will lead to significantly lower circuit aging than RENEU}.
}

\section{Conclusions}\label{sec:conclusions}
\rev{
This paper introduces \tech{}, a run-time reliability manager for neuromorphic computing. 
We observe that neurons and synapses in neuromorphic hardware are exposed to high voltages and/or currents because of the operating requirements of the Non-Volatile Memory, which are used for high density and low energy synaptic storage in the hardware. When exposed to these elevated conditions for too often, the CMOS transistors in the neuron and synapse circuit suffer strong aging, leading to hard breakdown. But in strongly scaled sub-10nm technology nodes, even under normal workloads,  parametric soft breakdown mechanisms will start drifting the transistor parameters from their nominal values.
In contrast to long-term aging, which permanently damages the hardware, short-term aging in scaled CMOS transistors is mostly due to BTI. The latter is heavily workload-dependent and more importantly, partially reversible. 
Based on these observations, \tech{} dynamically de-stresses neuron and synapse circuits in response to the short-term aging in their CMOS transistors during the execution of machine learning tasks, with the objective of meeting a reliability target. 
\tech{} de-stresses these circuits only when it is absolutely necessary to do so, otherwise reducing the performance impact by scheduling de-stress operations off the critical path. We evaluate \tech{} with supervised and unsupervised machine learning applications on a neuromorphic hardware. Our results demonstrate that
that for machine learning workloads with sparse activation, \tech{} is significantly better than design-time based approaches both in terms of reliability and performance. For dense activation, \tech{} improves reliability significantly compared to these approaches, with only marginal impact on performance. 
We conclude that \tech{} can be easily extended to incorporate other failure mechanisms.
}
\minorrev{
In the future, we plan on implementing \tech{} on NVM-based DYNAP-SE, when such board will be made publicly available.
}

\section{Acknowledgments}
This work is supported by the National Science Foundation Faculty Early Career Development Award CCF-1942697 (CAREER: Facilitating Dependable Neuromorphic Computing: Vision, Architecture, and Impact on Programmability).

\bibliographystyle{IEEEtranSN}
\bibliography{commands,disco,external}

\appendix

\section{Introduction to Spiking Neural Networks}\label{sec:snns}
\begin{figure}[h]
	\centering
	\centerline{\includegraphics[width=0.59\columnwidth]{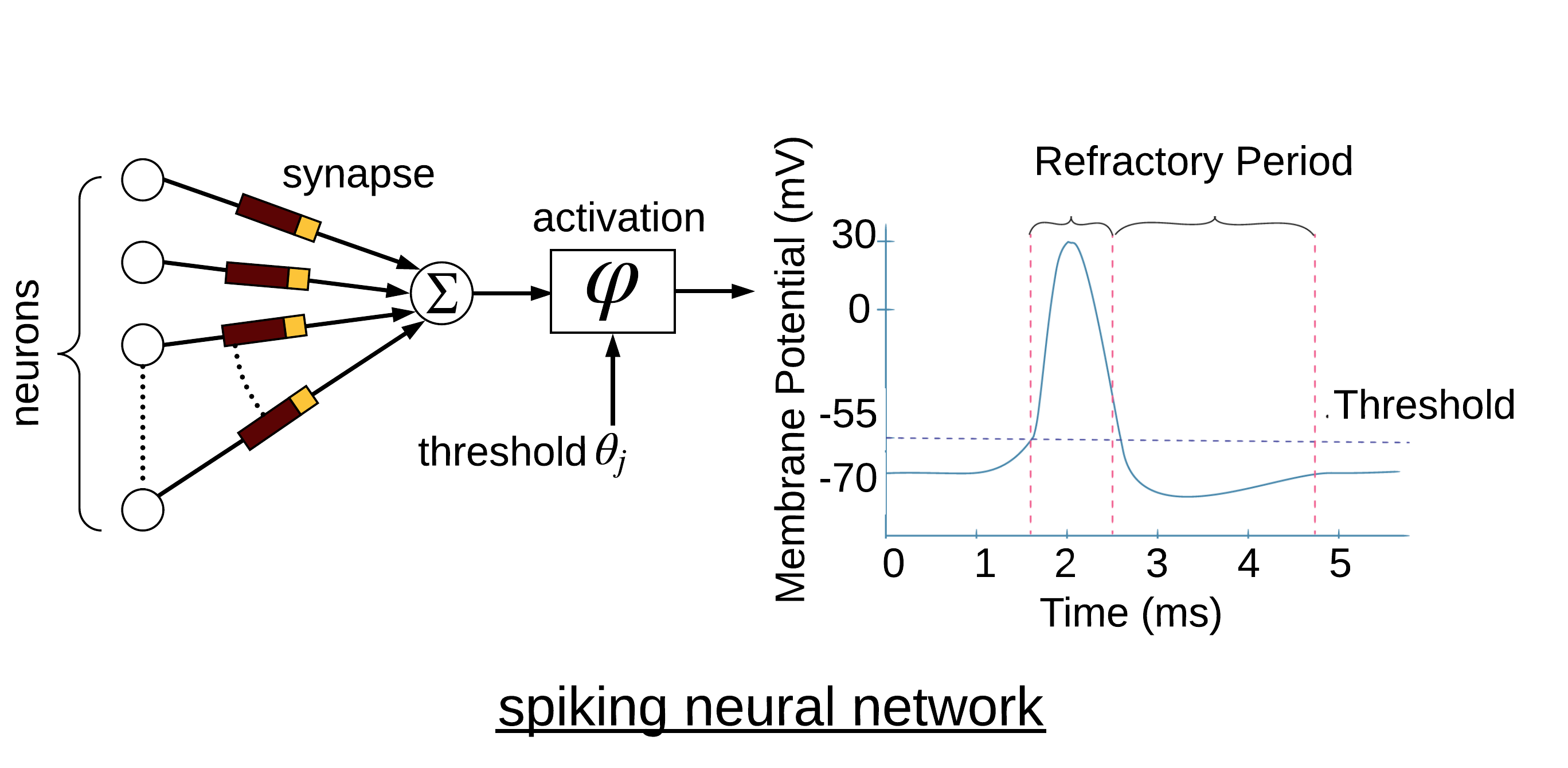}}
	\vspace{-10pt}
	\caption{Illustration of spiking neural networks.}
	\label{fig:snn}
\end{figure}

Spiking Neural Networks (SNNs)~\cite{maass1997networks} are regarded as the third generation of neural networks (see Figure~\ref{fig:snn}a). SNNs consist of spiking neurons, which are implemented using integrate and fire~\cite{Burkitt2006} model. 
In this model, a neuron fires a spike when its membrane voltage exceeds a threshold and subsequently the membrane voltage is reset. The moment of threshold crossing in a neuron defines its \textit{firing time}. Post firing, the neuron goes into a refractory state, where the neuron cannot be excited to generate a second action potential (no matter how intense the input stimulus be) (see Figure~\ref{fig:snn}b). Spiking neurons are interconnected via synapses as shown in Figure~\ref{fig:snn}a.

\textbf{Information Encoding in SNNs:} Information in SNNs can be encoded using different techniques \cite{schliebs2013evolving}, prominent among which are  rate coding~\cite{van2001rate} and temporal coding \cite{van2001novel}.
Rate coding encodes information as number of spikes within an encoding window without considering the temporal characteristics of the signal. Temporal coding encodes information as inter-spike interval (ISI), capturing the spatio-temopral structure of the input signal.

\textbf{Machine Learning Approaches using SNNs:} SNNs can be used to implement many machine learning approaches. One example is the supervised approach, where an SNN is first \textit{trained} with examples from the field and then used for \textit{inference} with current data. SNNs can also implement unsupervised, semi-supervised, and reinforcement learning-based machine learning approaches.

\textbf{Learning Algorithms in SNNs:} Currently, spike-based learning rules are limited, compared to the wide range of learning rules available for analog or rate-based artificial neural networks (ANNs). Most learning rules are based on unsupervised correlational learning rules, such as spike timing dependent plasticity (STDP)~\cite{Dan2004}, short-term plasticity (STP), and long-term plasticity (LTP). Other modifications include a localized version of backpropagation suitable for SNNs. However, these variants are supervisory and takes a while to converge. Attempts have been made to add a reinforcer to STDP based on the idea that dopamine in the brain carries a reward prediction error signal. In practice, dopamine modulated STDP (DA-STDP) takes a long time before the network has a strong enough signal to drive behavior~\cite{chou2015learning}. Recently, a reward-modulated STDP (R-STDP) learning is developed to train SNN controllers for obstacle avoiding behavior in mobile robots~\cite{legenstein2008learning}.

\section{Impact of ISI distortion on performance of SNNs}\label{sec:performance_loss}
\begin{figure}[h!]
	\centering
	\centerline{\includegraphics[width=0.59\columnwidth]{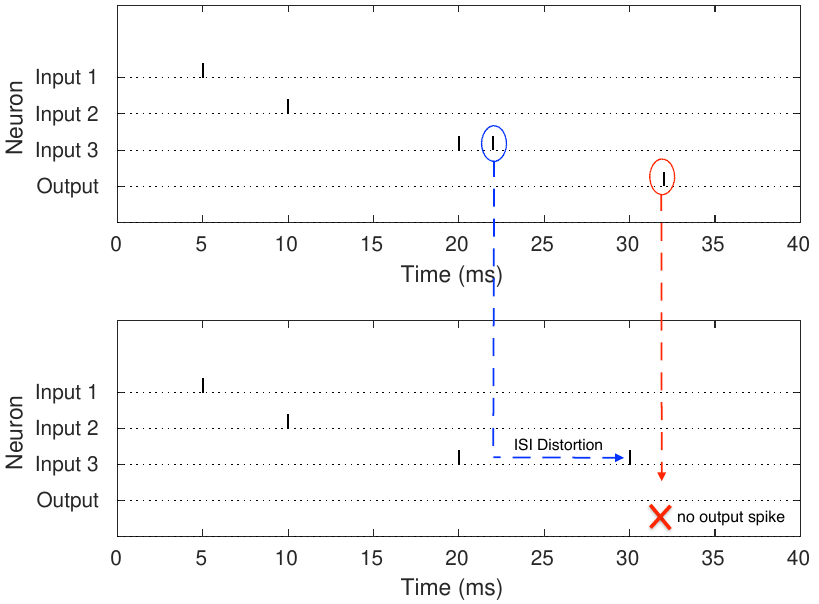}}
	\caption{Impact of ISI distortion on accuracy. Top sub-figure shows a scenario where an output spike is generated based on the spikes received from the three input neurons. Bottom sub-figure shows a scenario where the second spike from neuron 3 is delayed. As a result of this ISI distortion, there are no output spikes  generated. Loss of spikes can lead to accuracy drop.}
	\label{fig:isi_imact}
\end{figure}

\rev{
To illustrate how ISI distortion and spike disorder impact accuracy, we consider a small SNN example where three input neurons are connected to an output neuron. In Figure \ref{fig:isi_imact}, we illustrate the impact of ISI distortion on the output spike. In the top sub-figure, we observe that a spike is generated at the output neuron at 22ms due to spikes from the input neurons. In the bottom sub-figure, we observe that the second spike from input 3 is delayed, i.e., has ISI distortion. As a result of this distortion, there is no output spike. Missing spikes can impact application accuracy, as spike timings encode information in SNNs.
}

\end{document}